\documentclass{article}

\usepackage{microtype}
\usepackage{graphicx}
\usepackage{subfigure}
\usepackage{booktabs} 

\usepackage{hyperref}
\usepackage{url}

\usepackage[accepted]{icml2025}

\usepackage{amsmath}
\usepackage{amssymb}
\usepackage{mathtools}
\usepackage{amsthm}

\usepackage{amsmath,amsfonts,bm}

\def\eqref#1{equation~\ref{#1}}

\def\1{\bm{1}}

\def\vp{{\bm{p}}}

\def\vw{{\bm{w}}}
\def\vx{{\bm{x}}}

\def\mP{{\bm{P}}}

\def\mS{{\bm{S}}}

\def\mX{{\bm{X}}}

\DeclareMathAlphabet{\mathsfit}{\encodingdefault}{\sfdefault}{m}{sl}
\SetMathAlphabet{\mathsfit}{bold}{\encodingdefault}{\sfdefault}{bx}{n}

\usepackage{amsfonts}
\usepackage{nicefrac}       
\usepackage{xcolor}         
\usepackage{xspace}
\usepackage{multirow}
\usepackage{enumitem}
\usepackage{tabularx}
\usepackage{array}

\usepackage[capitalize,noabbrev]{cleveref}

\theoremstyle{plain}

\theoremstyle{definition}

\theoremstyle{remark}

\usepackage[textsize=tiny]{todonotes}

\newcommand{\eg}{\emph{e.g.}}
\newcommand{\ie}{\emph{i.e.}}

\graphicspath{{./}}

\newcommand{\tablestyle}[2]{\setlength{\tabcolsep}{#1}\renewcommand{\arraystretch}{#2}\centering\footnotesize}
\newcommand{\cmark}{\ding{51}}
\newcommand{\xmark}{\ding{55}}

\usepackage{pifont,makecell,subcaption}
\usepackage{wrapfig}
\usepackage{sidecap}
\usepackage{authblk}

\icmltitlerunning{Proxy-FDA}

\begin{document}

\twocolumn[
\icmltitle{Proxy-FDA: Proxy-based Feature Distribution Alignment \\ 
for Fine-tuning Vision Foundation Models without Forgetting}



\icmlsetsymbol{equal}{*}

\begin{icmlauthorlist}
\icmlauthor{Chen Huang}{comp}
\icmlauthor{Skyler Seto}{comp}
\icmlauthor{Hadi Pouransari}{comp}
\icmlauthor{Mehrdad Farajtabar}{comp}
\icmlauthor{Raviteja Vemulapalli}{comp}
\icmlauthor{Fartash Faghri}{comp}
\icmlauthor{Oncel Tuzel}{comp}
\icmlauthor{Barry-John Theobald}{comp}
\icmlauthor{Josh Susskind}{comp}
\end{icmlauthorlist}

\icmlaffiliation{comp}{Apple, Cupertino, United States}

\icmlcorrespondingauthor{Chen Huang}{chen-huang@apple.com}

\icmlkeywords{Proxy-FDA, robust fine-tuning, concept forgetting, vision foundation model}

\vskip 0.3in
]



\printAffiliationsAndNotice{}  

\begin{abstract}
Vision foundation models pre-trained on massive data encode rich representations of real-world concepts, which can be adapted to downstream tasks by fine-tuning. However, fine-tuning foundation models on one task often leads to the issue of \textit{concept forgetting} on other tasks. Recent methods of robust fine-tuning aim to mitigate forgetting of prior knowledge without affecting the fine-tuning performance. Knowledge is often preserved by matching the original and fine-tuned model weights or feature pairs. However, such point-wise matching can be too strong, without explicit awareness of the feature neighborhood structures that encode rich knowledge as well. We propose a novel regularization method \textbf{Proxy-FDA} that explicitly preserves the structural knowledge in feature space. Proxy-FDA performs Feature Distribution Alignment (using nearest neighbor graphs) between the pre-trained and fine-tuned feature spaces, and the alignment is further improved by informative proxies that are generated dynamically to increase data diversity. Experiments show that Proxy-FDA significantly reduces concept forgetting during fine-tuning, and we find a strong correlation between forgetting and a distributional distance metric (in comparison to L2 distance). We further demonstrate Proxy-FDA's benefits in various fine-tuning settings (end-to-end, few-shot and continual tuning) and across different tasks like image classification, captioning and VQA.
\end{abstract}

\section{Introduction}
Vision foundation models like CLIP~\citep{radford2021learning} and DINOv2~\citep{oquab2024dinov} pre-trained on large amounts of data demonstrate remarkable performance across various tasks and data distributions. Such foundation models are known to have learned vast knowledge on real-world concepts that can serve as a useful prior for downstream task adaptation via fine-tuning. Existing fine-tuning methods include end-to-end finetuning, linear probing, prompt tuning~\citep{zhou2021coop,zhou2022cocoop}, and adapter learning~\citep{gao2021clip}. While these methods prove effective, empirical evidence shows that they frequently suffer from an undesirable effect called \textit{concept forgetting}~\citep{mukhoti2024finetuning}. Forgetting occurs when a fine-tuned model overfits on the downstream task, and unlike its pre-trained counterpart, significantly loses the ability to recognize concepts on other tasks.

Concept forgetting has driven recent research on robust fine-tuning. The goal is to preserve the pre-trained knowledge~\emph{and}~perform well on downstream tasks. One simple approach is to ensemble models before and after fine-tuning~\citep{wortsman2022robust}. Alternative methods use regularization techniques to constrain the fine-tuned model to remain close to the original foundation model in either weight space~\citep{li18a} or feature space~\citep{mukhoti2024finetuning}.
Feature-space regularization by matching the pre-trained and fine-tuned features across samples shows a more promising effect in reducing forgetting, since it directly minimizes the change in input-output behaviour of the model. One key assumption behind such regularization is that the L2 feature-space distance is a good indicator of the similarity of encoded concepts in different models.

\begin{figure*}[!t]
\begin{center}
\vskip -0.1in
\centerline{\includegraphics[width=1.0\linewidth]{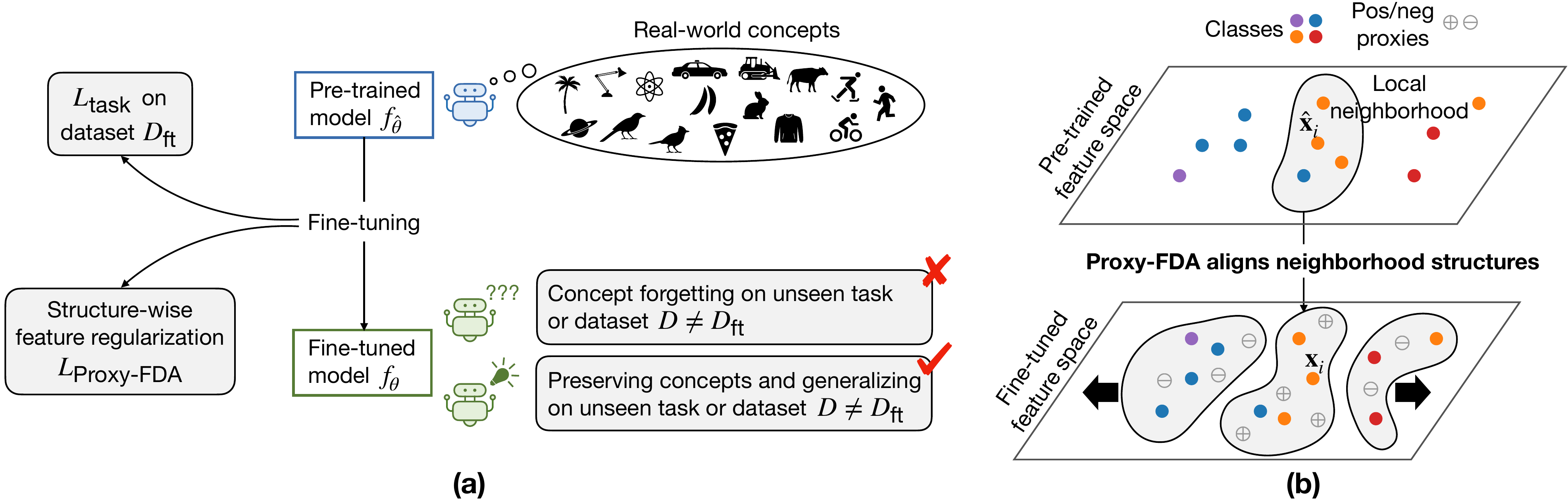}}
\end{center}
\vskip -0.35in
\caption{\textbf{(a)} Motivation: alleviating concept forgetting during model fine-tuning by a novel feature regularization method--Proxy-FDA (Proxy-based Feature Distribution Alignment). \textbf{(b)} Proxy-FDA aligns the pre-trained and fine-tuned feature distributions by their local neighborhood structures, which is further aided by proxies (\ie,~synthetic features). We show Proxy-FDA indeed preserves the rich knowledge in local feature neighborhood (example visualization in Fig.~\ref{fig:tsne}).} 
\label{fig:fig1}
\end{figure*}

We argue that aligning individual feature points imposes too strong of a constraint. Without an explicit insight of feature neighborhoods, the concepts preserved \textit{point-wise} are found to be limited, resulting in sub-optimal performance of forgetting reduction. Here we suggest that it is desirable to explicitly inform the fine-tuning process of the local structure of feature neighborhoods. By preserving this neighborhood structure with a \textit{structure-wise} regularization term, the rich knowledge encoded in the local structure of the original feature space will be transferred to the fine-tuned one. As a result, the fine-tuned model can forget significantly less while still maintaining its downstream performance (Fig.~\ref{fig:fig1}).

The above idea motivates us to propose a new feature regularization term called \textbf{Feature Distribution Alignment} (\textbf{FDA}). Specifically, we first model the structural relations of pre-trained features using a nearest neighbor graph. Then we transfer the graph to the fine-tuned feature space, where feature neighbors are pulled together while non-neighbors are pushed away regardless of their labels.
Such FDA process enables sharing knowledge beyond class concepts (\eg,~visual attributes) in local feature neighborhoods. Fig.~\ref{fig:tsne} provides an example of the white color attribute of two dog breeds mined from a local neighborhood on ImageNet.
This example represents the common-sense prior knowledge embedded in a vision foundation model that is often richer than the class labels on downstream datasets. Preserving such knowledge (\eg,~about color) during fine-tuning is important to maintain the generalizability of the foundation model, which can facilitate recognizing unfamiliar classes from different tasks. What is harmful is to just specialize on the task at hand, since all information (\eg,~color sensitivity) but its class label will be discarded.

Another key contribution of this paper is an improvement to FDA, with the introduction of a new regularization called \textbf{Proxy-FDA}, which uses \textit{proxies} as synthetic features. This full method is particularly useful on data-deficient fine-tuning tasks (such as few-shot ones), where the limited task data do not allow sufficient alignment of complex feature distributions.
To further increase data diversity, Proxy-FDA learns to generate a set of instance-wise proxies both within and outside a target feature's local neighborhood. Fig.~\ref{fig:tsne} exemplifies some proxies that synthesize informative unseen data or unseen class concepts. We empirically show that the generated proxies improves FDA with richer data/concepts, thereby further reducing concept forgetting.

We have extensive experiments of fine-tuning vision foundation models end-to-end on ten classification tasks. Results show that Proxy-FDA significantly outperforms other fine-tuning baselines in preventing concept forgetting, without hurting the downstream accuracy. We also find a strong correlation between concept forgetting and a distance metric OTDD -- Optimal Transport Dataset Distance~\citep{NEURIPS2020_f52a7b26} which is ideal to measure the alignment quality for feature distributions with local structures. Crucially, the correlation between concept forgetting and the OTDD metric indicates the need for some form of structure-wise FDA to better mitigate forgetting. Results confirm that our structure-wise Proxy-FDA forgets much less than point-wise feature regularization~\citep{mukhoti2024finetuning}.

We further show Proxy-FDA can be plugged into various prompt tuning methods for few-shot fine-tuning. In all cases, Proxy-FDA shows superior performance and data efficiency for lowering forgetting. Proxy-FDA also proves effective on continual fine-tuning tasks, outperforming specialized continual learning baselines. Lastly, we show the benefits of Proxy-FDA when fine-tuning for tasks beyond classification, like image captioning and VQA. Proxy-FDA also demonstrates its utility in the domain of knowledge distillation.

In summary, our \textbf{main contributions} include:
\vspace{-0.1in}
\begin{itemize}[leftmargin=20pt]
\setlength{\itemsep}{0pt}
\item A novel regularization method, Proxy-FDA, that aligns the local structures of feature distributions with learned proxies, aiming to preserve concepts when fine-tuning vision foundation models;
\item Correlation analysis between concept forgetting and a structure-aware distributional distance metric, OTDD, which implicitly explains the success of our structure-wise FDA method;
\item State-of-the-art performance on reducing forgetting in various fine-tuning settings and across different tasks.
\end{itemize}

\section{Related Work} 
\paragraph{Robust fine-tuning.} End-to-end fine-tuning often suffers from concept forgetting and degraded out-of-distribution (OOD) performance. In the foundation model era, linear probing or that followed by end-to-end tuning~\citep{kumar2022finetuning} are common remedies to maintain the OOD robustness of a pre-trained model. Alternative methods either ensemble the original and fine-tuned models~\citep{wortsman2022robust,wortsman22a} or use the contrastive pre-training loss directly for fine-tuning~\citep{10205046}. More recently, \citet{song2023fdalign} propose a method called FD-Align, which trains a spurious feature classifier and maintains its output consistency during fine-tuning. As a result, FD-Align significantly improves the OOD accuracy. To prevent forgetting, regularization methods are often used to minimize the model distance before and after fine-tuning in either weight space~\citep{li18a} or feature space~\citep{mukhoti2024finetuning}. In few-shot settings, regularization is even more important. For example, the prompt learning method CLIPood~\citep{shu2023clipood} regularizes via temporal model ensembling, while PromptSRC~\citep{khattak2023self} directly regularizes the output features and logits between pre-trained and prompt-tuned models.
Nevertheless, all existing methods do not explicitly account for feature neighborhood structures, which we show is key for robust fine-tuning.

\paragraph{Feature and data distribution alignment.} These techniques have been explored in different contexts. At the core of measuring \textbf{distributional distances}, Optimal Transport (OT)~\citep{villani2008optimal} provides a principled approach to compare data distributions in a geometrically meaningful way. Given the similar nature of our FDA method that aligns the ``clustering'' structures of distributions, we use an OT-based distance metric OTDD~\citep{NEURIPS2020_f52a7b26} to measure FDA quality. Feature alignment is also key to \textbf{Domain Adaptation} (DA)~\citep{DA_survey}. However, most DA methods learn a separate domain-invariant feature subspace to align domains implicitly, which differs from our explicit FDA during fine-tuning. More related to our method is the \textbf{Knowledge Distillation} (KD) field~\citep{Wang2020KnowledgeDA}, where traditional KD methods match features or probability distributions between teacher and student models. Relation-based KD methods are particularly similar to our high-level idea by distilling feature relations in form of kNNs~\citep{zhu2022CNA}, feature similarities~\citep{park2019relational,pkt_eccv,9010328,CCKD} and relative ranks~\citep{ChenWZ18}.
Our Proxy-FDA can be seen as an alternative relational KD method that distills both kNNs and similarities, and further improves with proxy learning.

\paragraph{Proxy learning.} This approach is widely adopted in deep metric learning~\citep{2017NoFD,Kim_2020_CVPR,RotVinAka22b} to reduce the sampling complexity of pure sample-based methods. Proxies are learned as class prototypes to optimize sample-proxy distances in place of sample-sample distances, resulting in faster convergence. By contrast, our proxy learning is different in both implementation and motivation: we learn instance-wise proxies via adaptive pooling of true samples; we also do not use the proxies as sample stand-ins, but as rich augmentations for improving FDA. This makes our approach more related to those \textbf{feature augmentation} methods, such as by random linear interpolation~\citep{verma19a} and outlier feature synthesis~\citep{du2022vos,tao2023nonparametric}. Empirically, our method is more effective than these feature augmentation methods by generating diverse augmented features from the entire feature neighborhood. 

\section{Method}

We aim for forgetting-free fine-tuning of vision foundation models (\eg,~CLIP and DINOv2), using feature-space regularization based on \textit{Feature Distribution Alignment} (FDA). Specifically, given a pre-trained model $f_{\hat\theta}$, we use the downstream dataset $\mathcal{D}_{\text{ft}}$ to fine-tune the model into $f_\theta$. Our goal is to specialize the fine-tuned model on $\mathcal{D}_{\text{ft}}$ with low task loss $\mathcal{L}_{\text{task}}$ (\eg,~cross-entropy loss for classification), whilst preventing concept forgetting on any target dataset $\mathcal{D} \neq \mathcal{D}_{\text{ft}}$. To prevent forgetting, we introduce an FDA-based regularization term to the downstream task loss, which gives:
\begin{equation}
    \mathcal{L}=\frac{1}{B} \sum_{i=1}^{B} \left( \mathcal{L}_{\text{task}}^i+\lambda \mathcal{L}_{\text{FDA}}^i \right),
\label{eq1}
\end{equation}
where $\mathcal{L}_{\text{FDA}}^i$ is the FDA loss for each sample $i$ in a mini-batch $\{i\}_{i=1}^B$ of size $B$, and $\lambda$ is a weighting parameter.

\begin{figure*}[!t]
\begin{center}
\vskip -0.1in
\centerline{\includegraphics[width=0.85\linewidth]{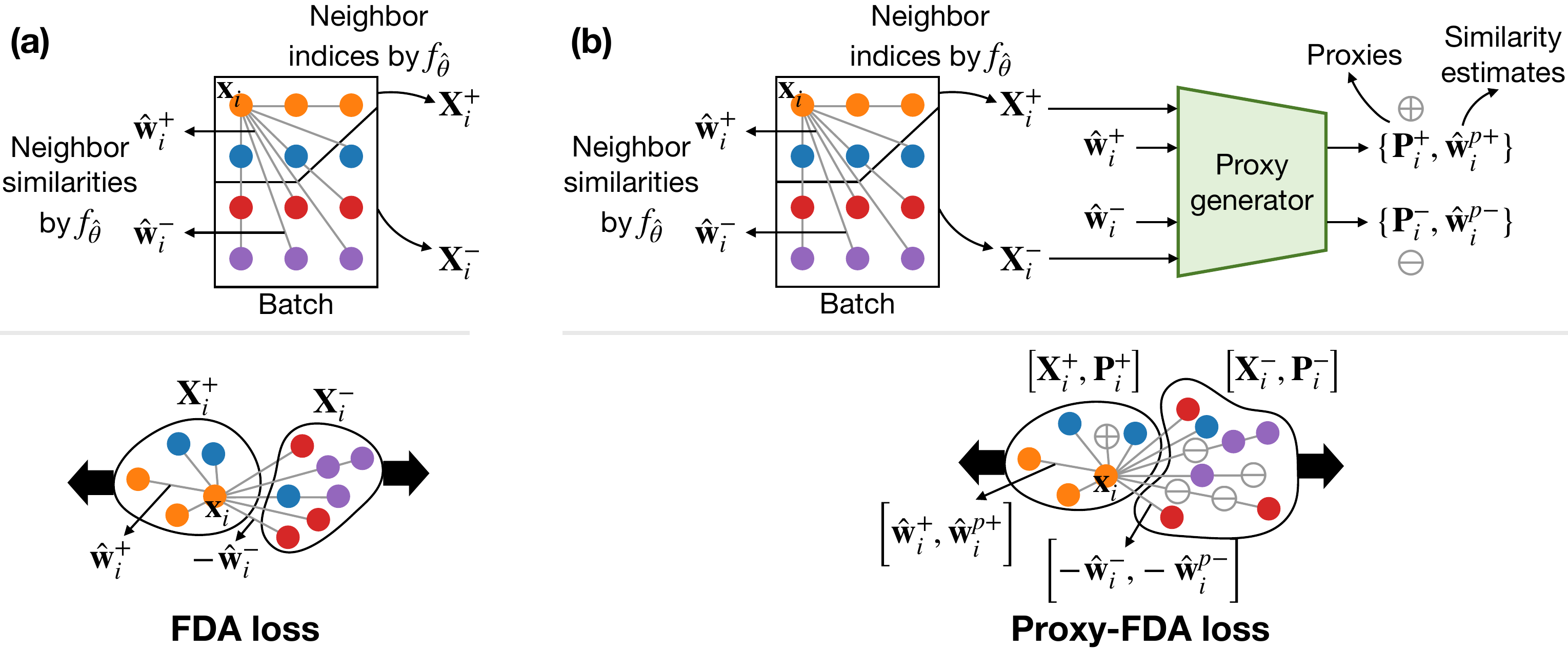}}
\end{center}
\vskip -0.35in
\caption{Batch construction and nearest neighbor graph transfer for our \textbf{(a)} FDA and \textbf{(b)} Proxy-FDA methods, both regularizing the fine-tuned features $\{\mX_{i}^{+},\mX_{i}^{-}\}$ using the pre-trained model $f_{\hat\theta}$. (Proxy-) FDA loss penalizes local distribution overlap between the positives $\mX_{i}^{+}$ and negatives $\mX_{i}^{-}$ (weighted by the associated similarities $\hat\vw_{i}^{+}$ and $\hat\vw_{i}^{-}$), (with) and without using the generated proxies $\{\mP_{i}^{+},\mP_{i}^{-}\}$ and their similarity estimates $\{\hat\vw_{i}^{p+},\hat\vw_{i}^{p-}\}$. Fig.~\ref{fig:proxy_generator} shows the network architecture of our proxy generator (details in Appendix~\ref{sec:proxy_generator}).}
\label{fig:schema_backup}
\vskip -0.18in
\end{figure*}

\subsection{Feature Distribution Alignment (FDA)}
\label{sec:fda_baseline}

Having defined the learning problem and its general loss function, we now present our FDA method in detail. During fine-tuning on $\mathcal{D}_{\text{ft}}$, we first use the pre-trained model $f_{\hat\theta}$ and fine-tuned $f_{\theta}$ to extract batch features $\hat\mX \in \mathbb{R}^{d \times B}$ and $\mX \in \mathbb{R}^{d \times B}$, respectively. Note $\hat\mX=[\hat\vx_{1},\dots,\hat\vx_{B}]$ are the pre-trained batch features with $\hat\vx_{i} \in \mathbb{R}^d$, while $\mX=[\vx_1,\dots,\vx_B]$ are the features currently being fine-tuned with $\vx_{i} \in \mathbb{R}^d$. To transfer the structural knowledge in $\hat\mX$ into $\mX$, we align the structural relations of $\hat\mX$ and $\mX$ based on their nearest neighbor graphs.

Concretely, for each pre-trained feature point $\hat\vx_{i}$, we maintain its $k$-nearest neighbor set $R_i=\{j|\hat\vx_{j} \in \text{kNN}(\hat\vx_{i})\}$ within the batch. Note $|R_i|=K$, and we will detail later how to construct batches to facilitate the kNN search. This way, we obtain an instance-wise batch partition from the pre-trained model's perspective, leading to the positive set of neighbors $\hat\mX_{i}^{+}=\hat\mX(R_i) \in \mathbb{R}^{d \times K}$ and negative set of non-neighbors $\hat\mX_{i}^{-} \in \mathbb{R}^{d \times (B-K-1)}$.
To form the complete nearest neighbor graph, we further compute the cosine similarities between pre-trained features $\hat w_{ij} = \cos(\hat\vx_{i},\hat\vx_{j})$ for $j\in \{1,\dots,B\}$ and $j\neq i$. Accordingly, we organize them into similarity vectors for neighbors $\hat\vw_{i}^{+} \in \mathbb{R}^{K}$ and non-neighbors $\hat\vw_{i}^{-} \in \mathbb{R}^{B-K-1}$.


For efficient graph matching between $\hat\mX$ and $\mX$ , we choose to simply transfer the neighbor indices $R_i$ and similarities $\{\hat\vw_{i}^{+},\hat\vw_{i}^{-}\}$ from $\hat\mX$ to $\mX$.
This means neighbors in the pre-trained feature space should remain neighbors in the fine-tuned feature space. Hence among $\mX$, we similarly have a positive set $\mX_{i}^{+}=\mX(R_i) \in \mathbb{R}^{d \times K}$ where the identified neighbors are pulled together in the fine-tuned feature space, and a negative set $\mX_{i}^{-} \in \mathbb{R}^{d \times (B-K-1)}$ where non-neighbors are pushed away. On the other hand, we associate the pre-trained feature similarities $\{\hat\vw_{i}^{+},\hat\vw_{i}^{-}\}$ with $\{\mX_{i}^{+},\mX_{i}^{-}\}$ to preserve fine-grained feature neighborhood structures. We will show that transferring both the neighbor indices and similarities works better than only transferring neighbor indices. Fig.~\ref{fig:schema_backup}(a) visualizes the high-level idea.

To capture the desired structures, we use the noise-resistant Sigmoid loss~\citep{Zhai_2023_ICCV} to handle a variable number of positives and negatives per batch:
\begin{eqnarray}
    \!\!\!\!\!\!\!\!\!\!\!\!\!\! &&\mathcal{L}_{\text{FDA}}^i\left(\{\mX_{i}^{+},\mX_{i}^{-}\},\{\hat\vw_{i}^{+},\hat\vw_{i}^{-}\}\right)= \\ \nonumber
    \!\!\!\!\!\!\!\!\!\!\!\!\!\! && \frac{1}{(|\mX|-1)} \sum_{\vx_{j}\in \mX, j\neq i} \!\!\!\! \log \left(  1+ e^{w_{ij} \left(-\frac{\cos(\vx_{i},\vx_{j})}{\tau}+b \right)} \right),
\label{eq:fda}
\end{eqnarray}
where $w_{ij}$ is a weighting parameter. $w_{ij}$ equals $\hat w_{ij}$ if $j\in R_{i}$ (\ie,~weighting by $\hat\vw_{i}^{+}$ for neighbors), and $-\hat w_{ij}$ if $j\notin R_{i}$ (\ie,~weighting by $-\hat\vw_{i}^{-}$ for non-neighbors). $\tau$ and $b$ are learnable parameters which are initialized in a similar way as in~\citep{Zhai_2023_ICCV}. The above FDA loss helps to preserve local neighborhood structures in the fine-tuned feature space, without involving class labels.

\textbf{Batch construction and neighborhood size $K$}. To have a meaningful characterization \textit{and} alignment of local neighborhood structures, we need to ensure that each mini-batch has diverse class distributions that may overlap locally in the feature space, and that a sufficient number of neighbors $|R_i|=K$ are identified for each sample in a batch.

To meet the above requirements, we \textit{sample batch data in a class-balanced manner}, with $n$ samples for each of the $m$ classes. For a fixed batch size $B=m\cdot n$ that best fits in the available GPU memory, we choose a high value of $m$ to increase the diversity of class concepts in batch, but at the cost of reducing the number of examples per class $n$. By default, $m=16$ and $n=4$. More critically, we perform \textbf{hard class mining} to construct batches where samples from different classes are similar (details in Appendix~\ref{sec:hard_mining}). This enables meaningful kNN search within a batch.

\textit{For the neighborhood size} $K$, we choose $K>n$ to guarantee that there is more than one class in \textit{any} identified local feature neighborhood $R_i$. This way, each neighborhood includes an adaptive selection of ``small clusters'' from related classes. FDA between such neighborhoods will encourage transferring high-level knowledge beyond class concepts. Fig.~\ref{fig:tsne} exemplifies a common color attribute mined for two similar dog classes. Preserving this knowledge that is embedded in foundation models is important to prevent forgetting during fine-tuning. Note it is possible that the inter-class similarity is not high enough in $R_i$ (thus relatively low $\hat w_{ij}$ for inter-class samples and there are no shared properties between neighboring classes). In this case, FDA reverts back to aligning class semantics.

\subsection{Proxy-FDA}

One challenge with FDA is that the downstream dataset $\mathcal{D}_{\text{ft}}$ can be limited in both data size and diversity. In this case $\mathcal{D}_{\text{ft}}$ does not allow adequate FDA, thereby preserving only limited concepts from those learned during pre-training. To address the data challenge, one could retrieve external data. However, using external data will inevitably suffer from higher compute/memory cost as well as various levels of distributional shift.
Here we propose a compute- and data-efficient approach to improve downstream data diversity and eventually improve FDA quality. Our approach involves generating synthetic features or \textit{proxies} on-the-fly from observed fine-tuning data. Such generated proxies have no distributional shift since they adapt to the considered feature distribution. We leave sampling suitable external data for FDA to future work. 


For proxy synthesis, we learn to generate two-sets of proxies $\mP_{i}^{+}=[\vp_1^+,\dots,\vp_{n^{p+}}^+] \in \mathbb{R}^{d \times n^{p+}}$ and $\mP_{i}^{-} = [\vp_1^-,\dots,\vp_{n^{p-}}^-] \in \mathbb{R}^{d \times n^{p-}}$ out of $\mX_{i}^{+} \in \mathbb{R}^{d \times K}$ and $\mX_{i}^{-} \in \mathbb{R}^{d \times (B-K-1)}$ respectively. $n^{p+}$ and $n^{p-}$ are made proportional to the size of $\mX_{i}^{+}$ and $\mX_{i}^{-}$ using a scalar $s$, see details in Appendix~\ref{sec:hyper-parameters}. The proxies are learned to be as diverse as possible but still lie in the corresponding true feature manifold.
Fig.~\ref{fig:tsne} shows that both $\mP_{i}^{+}$ and $\mP_{i}^{-}$ can synthesize unseen data/concepts. Such unseen information will provide fine-grained regularization of the neighborhood boundary, and will improve FDA with richer concepts.

Following the above intuitions, we define our proxy learning loss $\mathcal{L}_{\text{proxy}}^{i}=\mathcal{L}_{\mP_i^+}+\mathcal{L}_{\mP_i^-}$, where:
\begin{eqnarray}
\!\!\!\!\!\!\!\!\!\!\mathcal{L}_{\mP_i^+}\!\!\!\!\!\!\!\!\!\!\!\! &&=\!\!\frac{1}{n^{p+}}\sum_{j=1}^{n^{p+}}\frac{1}{|\mX|} \sum_{\vx_{l}\in \mX} \!\!\log \left(  1+ e^{w_{l} \left(-\frac{\cos(\vp_{j}^+,\vx_{l})}{\tau}+b \right)} \right) \\ \notag
\!\!\!\!\!\!\!\!\! && \;\;\; + \alpha \cdot \mathcal{L}_{\text{var}}(\mP_i^+), \\
\!\!\!\!\!\!\!\!\!\!\mathcal{L}_{\mP_i^-}\!\!\!\!\!\!\!\!\!\!\!\! &&=\!\!\frac{1}{n^{p-}}\sum_{j=1}^{n^{p-}}\frac{1}{|\mX|} \sum_{\vx_{l}\in \mX} \!\!\log \left(  1+ e^{w_{l} \left(-\frac{\cos(\vp_{j}^-,\vx_{l})}{\tau}+b \right)} \right) \\ \notag
\!\!\!\!\!\!\!\!\! && \;\;\; + \alpha \cdot \mathcal{L}_{\text{var}}(\mP_i^-).
\label{eq_proxy}
\end{eqnarray}
The first loss term constrains proxies $\mP_{i}^{+}$ and $\mP_{i}^{-}$ towards the feature manifolds $\mX_{i}^{+}$ and $\mX_{i}^{-}$. This is achieved using the binary label $w_{l}$ which, in case of $\mathcal{L}_{\mP_i^+}$, equals 1 if $\vx_{l}\in \mX_{i}^{+}$ and -1 if $\vx_{l}\in \mX_{i}^{-}$; while in case of $\mathcal{L}_{\mP_i^-}$, is the opposite. The variance loss $\mathcal{L}_{\text{var}}(\mP)$ maximizes proxy diversity in form of $1/d \sum_{j=1}^{d} \max(0,1-\sqrt{\text{Var}(\mP_{j,:})+\epsilon})$ with $\epsilon$ being a small scalar. $\alpha$ is a weighting parameter.

In practice, we use Eq.~(3-4) to train our proxy generator online during the model fine-tuning process. This ensures the generated proxies always adapt to the current feature distribution. Fig.~\ref{fig:proxy_generator} and Appendix~\ref{sec:proxy_generator} detail the \textbf{network architecture of the proxy generator}. At high level, conditioned on $\mX_{i}^{+}$ and $\mX_{i}^{-}$, our proxy generator is trained to predict the instance-wise proxies $\{\mP_{i}^{+},\mP_{i}^{-}\}$ and their similarity estimates $\{\hat\vw_{i}^{p+},\hat\vw_{i}^{p-}\}$ all at once. Finally, we use all the predictions to augment the true features $\{\mX_{i}^{+},\mX_{i}^{-}\}$ and similarities $\{\hat\vw_{i}^{+},\hat\vw_{i}^{-}\}$, arriving at our Proxy-FDA loss for feature-space regularization (see also Fig.~\ref{fig:schema_backup}(b)):
\begin{eqnarray}
    \mathcal{L}_{\text{Proxy-FDA}}^i \!\!\!\!\!\!\!\! &&  = \mathcal{L}_{\text{FDA}}^i\left(\left\{[\mX_{i}^{+},\mP_{i}^{+}],[\mX_{i}^{-},\mP_{i}^{-}]\right\},\right. \notag\\
    &&  \left. \left\{[\hat\vw_{i}^{+},\hat\vw_{i}^{p+}],[\hat\vw_{i}^{-},\hat\vw_{i}^{p-}]\right\}\right), \notag\\
    \mathcal{L} \!\!\!\!\!\!\!\! &&  =\frac{1}{B} \sum_{i=1}^{B} \left( \mathcal{L}_{\text{task}}^i+\lambda \mathcal{L}_{\text{Proxy-FDA}}^i \right).
\label{eq:proxyfda}
\end{eqnarray}
\vskip -0.1in

\section{Experiments}

In this section, we benchmark concept forgetting and different methods in 3 settings: end-to-end, few-shot and continual fine-tuning for image classification. We then move on to fine-tuning tasks of image captioning and VQA, and lastly to the application to knowledge distillation.
Appendix~\ref{sec:hyper-parameters} studies the hyper-parameters of our Proxy-FDA method, and Appendix~\ref{sec:ablating_proxyfda} ablates the key components of Proxy-FDA.

\textbf{Compute cost.} Our Proxy-FDA mainly involves FDA and proxy generation. The proxy generator is lightweight with only one attention and two convolutional layers (totalling 23.6k parameters), which is negligible in comparison to the foundation model size.
Here we show our feature regularization process only incurs a decent compute cost (on Nvidia A100 GPU). For end-to-end and few-shot fine-tuning tasks, averaged across the corresponding datasets, Proxy-FDA increases the fine-tuning time by 17\% and 21\% respectively, while FDA increases by 7\% and 9\%. Note Proxy-FDA does not impact the inference stage, hence we maintain the same FPS at the test time. 

\begin{table*}[!t]
\vskip -0.05in
\caption{\textbf{Test accuracy $\mathcal{A}_{\text{LP}}$ of end-to-end fine-tuned model on each dataset and its average $\Delta_{\text{LP}}$ computed over other datasets}.
The image encoder of CLIP ViT-B/32 is used here.
$\Delta_{\text{LP}}$ denotes the change in $\mathcal{A}_{\text{LP}}$ between pre-trained and fine-tuned models on target dataset $\mathcal{D}$, quantifying the level of concept forgetting. Higher $\Delta_{\text{LP}}$ shows lower forgetting or positive forward transfer ($\Delta_{\text{LP}}>0$).}
\vskip -0.2in
\label{tb:e2e_full}
\begin{center}
\resizebox{0.8\linewidth}{!}{
\begin{tabular}{l|cc|cc|cc|cc|cc|cc}
\toprule
\multicolumn{1}{l@{}}{Dataset} & \multicolumn{2}{c}{Naive End-to-End} & \multicolumn{2}{c}{LP-FT} & \multicolumn{2}{c}{L2SP} & \multicolumn{2}{c}{LDIFS} & \multicolumn{2}{c}{FDA (ours)} & \multicolumn{2}{c}{Proxy-FDA (ours)} \\
\cmidrule(lr){2-3} \cmidrule(lr){4-5} \cmidrule(lr){6-7} \cmidrule(lr){8-9} \cmidrule(lr){10-11} \cmidrule(lr){12-13}
\multicolumn{1}{c@{}}{} & $\mathcal{A}_{\text{LP}}$ & \multicolumn{1}{c@{}}{$\Delta_{\text{LP}}\uparrow$}& $\mathcal{A}_{\text{LP}}$ & \multicolumn{1}{c@{}}{$\Delta_{\text{LP}}\uparrow$} & $\mathcal{A}_{\text{LP}}$ & \multicolumn{1}{c@{}}{$\Delta_{\text{LP}}\uparrow$} & $\mathcal{A}_{\text{LP}}$ & \multicolumn{1}{c@{}}{$\Delta_{\text{LP}}\uparrow$} & $\mathcal{A}_{\text{LP}}$ & \multicolumn{1}{c@{}}{$\Delta_{\text{LP}}\uparrow$} & $\mathcal{A}_{\text{LP}}$ & $\Delta_{\text{LP}} \uparrow$\\
\midrule
Cars   & 83.48 &-1.56 &84.95 &-0.63 &83.87 &0.47 &85.26 &-0.18 &\textbf{85.36} &1.02 &84.69 &\textbf{1.26}\\
CIFAR10  & \textbf{97.73} &-1.60 &97.71 &-0.81 &97.66 &1.16 &97.24 &1.18 &97.53 &1.55 &97.61 &\textbf{1.63}\\
CIFAR100 & 88.60 &-0.96 &88.41 &-0.11 &86.94 &1.03 &\textbf{88.99} &0.86 &88.21 &1.44 &88.33 &\textbf{1.51}\\
DTD & 77.18 &-3.01 &72.18 &-1.76 &74.63 &0.01 &75.27 &0.53 &77.22 &1.04 &\textbf{77.28} &\textbf{1.19} \\
EuroSAT & 98.76 & -5.72 & \textbf{98.87} &-3.75 &98.20 & -0.85& 98.22 &1.32 &98.53 &1.61 &98.63 &\textbf{1.74} \\
GTSRB &98.52 &-5.90 & \textbf{98.53} & -0.94 &95.00 & 1.18 & 97.81 &1.27 &98.16 &1.58 &97.79 &\textbf{1.69} \\
MNIST &99.67 &-8.76 &\textbf{99.68} &-6.02 &99.18 &1.49 &99.52 &2.64 &99.43 &2.76 &99.49 & \textbf{2.81} \\
RESISC45 &\textbf{95.76} & -3.79 & 95.56 & -2.27 & 94.13 &0.66 &95.13& 0.90 & 95.31 &1.18 & 95.63 &\textbf{1.43} \\
SVHN &97.30 & -11.12 &\textbf{97.50} & -8.73 &96.54 &-2.11 &96.95 &-0.29 &96.96 &0.67 &96.65 &\textbf{0.92} \\
ImageNet &82.02 &-1.26 &82.12 &-0.87 &80.78 &-0.10 & \textbf{82.21} & 0.35 & 81.93 & 1.05 & 82.16 & \textbf{1.22} \\ \midrule
Mean across &\multirow{2}{*}{\textbf{91.90}} &\multirow{2}{*}{-4.37} &\multirow{2}{*}{91.55} &\multirow{2}{*}{-2.59} &\multirow{2}{*}{90.69} &\multirow{2}{*}{0.29} & \multirow{2}{*}{91.66} & \multirow{2}{*}{0.86} & \multirow{2}{*}{91.86} & \multirow{2}{*}{1.39} & \multirow{2}{*}{91.82} & \multirow{2}{*}{\textbf{1.54}} \\
10 datasets &  &  &  &  &  &  &   &   &   &  &   &   \\
\bottomrule
\end{tabular}
}
\end{center}
\end{table*}

\begin{figure*}[!t]
\begin{center}
\centerline{\includegraphics[width=1.0\linewidth]{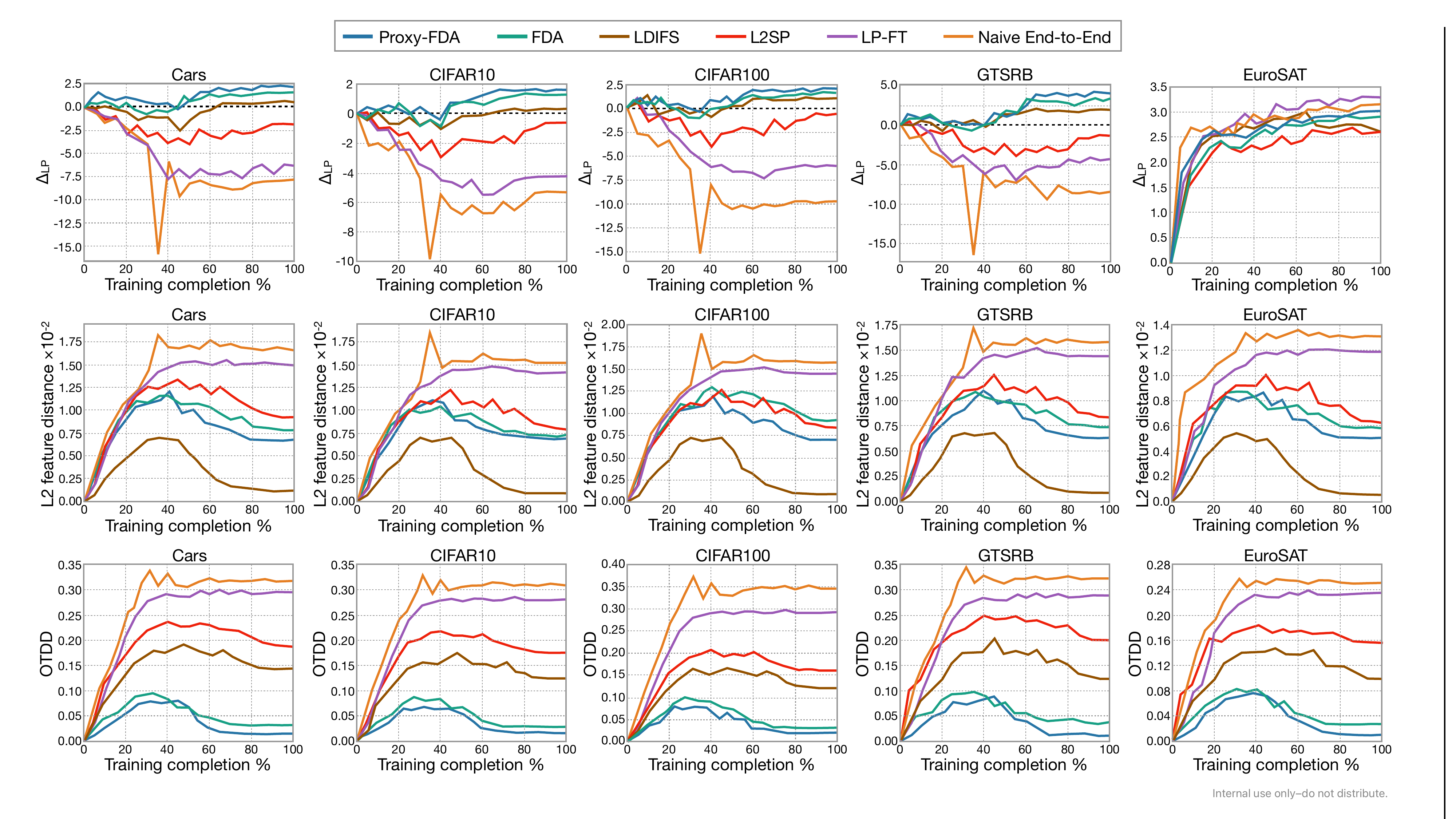}}
\end{center}
\vskip -0.3in
\caption{Three metrics computed over the course of model fine-tuning (CLIP ViT-B/32) on \textbf{EuroSAT}: $\Delta_{\text{LP}}$ (\textbf{Top row}), L2 feature-space distance (\textbf{Middle row}) and distributional distance metric OTDD (\textbf{Bottom row}), all between pre-trained and fine-tuned models. Our (Proxy-)FDA achieves the best results in preventing concept forgetting on other datasets (highest positive $\Delta_{\text{LP}}$) without hurting the downstream performance on EuroSAT. We also observe that concept forgetting measured by $\Delta_{\text{LP}}$ is more correlated to OTDD than L2 feature distance (see text for details).}
\label{fig:forget_OTDD}
\end{figure*}

\subsection{End-to-End Fine-tuning}

\paragraph{Datasets.} We follow~\citep{mukhoti2024finetuning} to use 10 image classification datasets: Stanford Cars~\citep{krause20133d}, CIFAR-10/100~\citep{Krizhevsky09learningmultiple}, DTD~\citep{cimpoi2014describing}, EuroSAT~\citep{helber2019eurosat}, GTSRB~\citep{StallkampSSI12}, MNIST~\citep{lecun2010mnist}, RESISC45~\citep{RESISC45}, SVHN~\citep{SVHN} and ImageNet~\citep{deng2009imagenet}. These datasets include various semantic concepts, thus are perfect to benchmark forgetting of the rich pre-trained concepts.

\paragraph{Setting and baselines.} The image encoder of CLIP model (ViT-B/32) is fine-tuned end-to-end on the 10 datasets. We compare with popular end-to-end fine-tuning methods all using the cross-entropy loss as $\mathcal{L}_{\text{task}}$. The baselines include naive fine-tuning and LP-FT methods~\citep{kumar2022finetuning}. They differ in the linear head initialization, with zero-shot weights (text encodings of class name) and Linear Probe (LP) weights respectively. While L2SP~\citep{li18a} and LDIFS~\citep{mukhoti2024finetuning} add a point-wise regularization between the original and fine-tuned models in weight- and feature-space respectively. By contrast, our (Proxy-)FDA imposes a structure-wise regularization in feature space.
Note except for the naive fine-tuning baseline, LP initialization is used for all methods including ours for a fair comparison of different regularization techniques.

\paragraph{Metrics.} When fine-tuning on dataset $\mathcal{D}_{\text{ft}}$, we report two evaluation metrics: LP accuracy $\mathcal{A}_{\text{LP}}$ on the test set of $\mathcal{D}_{\text{ft}}$ (\ie,~the fine-tuning performance itself), and the change $\Delta_{\text{LP}}$ in $\mathcal{A}_{\text{LP}}$ between pre-trained and fine-tuned models on a different dataset $\mathcal{D}\neq \mathcal{D}_{\text{ft}}$. Negative $\Delta_{\text{LP}}$ indicates \textbf{concept forgetting} on $\mathcal{D}$, while a positive value indicates \textbf{positive forward transfer}. Clearly, the higher $\Delta_{\text{LP}}$ the better. When $\mathcal{D}= \mathcal{D}_{\text{ft}}$, $\Delta_{\text{LP}}$ on $\mathcal{D}$ simply denotes the change of downstream performance, and we expect $\Delta_{\text{LP}}$ to increase over the course of fine-tuning.

To gain insights on what impacts the concept forgetting performance, we further monitor two distance metrics for distribution alignment during fine-tuning: point-wise L2 distance between pre-trained and fine-tuned feature pairs, and Optimal Transport Dataset Distance (OTDD)~\citep{NEURIPS2020_f52a7b26} that takes feature distribution structures into consideration (details in Appendix~\ref{sec:otdd}). Between the two distance metrics, OTDD is generally more suited to measure the alignment quality for feature distributions with local structures as in our case.

\paragraph{Results.} Table~\ref{tb:e2e_full} compares $\mathcal{A}_{\text{LP}}$ on each fine-tuning dataset and the $\Delta_{\text{LP}}$ averaged over other datasets. We observe that FDA obtains a positive average $\Delta_{\text{LP}}$ for all fine-tuning tasks, thereby achieving a positive forward transfer. Proxy-FDA further improves the average $\Delta_{\text{LP}}$ consistently. This is not the case for naive fine-tuning and LP-FT where the average $\Delta_{\text{LP}}$ is all negative indicating concept forgetting. Point-wise regularization methods L2SP and LDIFS obtain mostly positive $\Delta_{\text{LP}}$ but significantly lower than our results, highlighting the benefits of our structure-wise feature regularization and proxy feature generation.

We also observe that our good performance on forgetting prevention does not compromise (much) the downstream fine-tuning accuracy $\mathcal{A}_{\text{LP}}$. The mean $\mathcal{A}_{\text{LP}}$ (across 10 datasets) of (Proxy-)FDA is (91.82) 91.86, which is only slightly lower than that of naive fine-tuning 91.90 but outperforms all other results. Overall, our structure-wise regularization method achieves the best trade-off between concept forgetting and downstream performance.
Fig.~\ref{fig:forget_OTDD} (top row) exemplifies the fine-tuning task on EuroSAT, where (Proxy-)FDA consistently outperforms other baselines in forgetting prevention during fine-tuning (higher $\Delta_{\text{LP}}$), but has similar performance on EuroSAT in the meantime.

Fig.~\ref{fig:forget_OTDD} (middle and bottom rows) shows how L2 feature distance and OTDD change during EuroSAT fine-tuning. Overall, both the distance metrics are correlated to concept forgetting --- fine-tuning methods with smaller L2 distance/OTDD forget less with higher $\Delta_{\text{LP}}$, while methods with a larger distance suffer more from forgetting with lower $\Delta_{\text{LP}}$. The only exception to the overall trend is when we use L2 feature distance to compare (Proxy-)FDA with LDIFS.
We see that (Proxy-)FDA, while having larger L2 distance than LDIFS, still forgets less. On the contrary, (Proxy-)FDA consistently gets lower OTDD. This suggests that \textbf{the structure-aware OTDD is a better indicator of concept forgetting compared to the point-wise L2 distance}. More crucially, the fact that OTDD is more correlated to forgetting than L2 distance reaffirms that having some form of structure-wise FDA can mitigate forgetting better.
Finally, we note our (Proxy-)FDA is only applied on EuroSAT samples, but the mitigation of forgetting extends to all other datasets. This indicates the generalizing effect of our feature regularization method, which can preserve pre-trained knowledge without requiring third party datasets during fine-tuning.


Table~\ref{tb:e2e_across_architectures} in Appendix shows that our benefits still hold when end-to-end fine-tuning happens with different architectures of CLIP~\citep{radford2021learning}, FLAVA~\citep{singh2022flava}, DINOv2~\citep{oquab2024dinov} and MAE~\citep{He2022}. Proxy-FDA consistently provides the highest $\Delta_{\text{LP}}$ values across foundation models and architectures, achieving positive forward transfer in all cases. Proxy-FDA also achieves the best $\mathcal{A}_{\text{LP}}$ in many cases, which is encouraging.

\subsection{Few-shot Prompt Tuning}

\paragraph{Datasets.} We follow~\citep{zhou2022cocoop} to use 11 datasets, consisting of a wide range of visual concepts again: ImageNet~\citep{deng2009imagenet}, Caltech101~\citep{fei2004learning}, OxfordPets~\citep{parkhi2012cats}, StanfordCars~\citep{krause20133d}, Flowers102~\citep{nilsback2008automated}, Food101~\citep{bossard2014food}, FGVC-Aircraft~\citep{maji2013fine}, SUN397~\citep{xiao2010sun}, DTD~\citep{cimpoi2014describing}, EuroSAT~\citep{helber2019eurosat} and UCF101~\citep{soomro2012ucf101}.

\begin{table*}[t!]
\centering
\caption{\textbf{Few-shot prompt tuning in the base-to-new class generalization setting} (16 shots per class). $\mathcal{A}_{\text{H}}$ denotes the Harmonic mean of $\mathcal{A}_{\text{Base}}$ and $\mathcal{A}_{\text{New}}$. $\Delta_{\text{New}}$ denotes the change in $\mathcal{A}_{\text{New}}$ between pre-trained and prompt-tuned CLIP models. Higher $\Delta_{\text{New}}$ shows lower level of concept forgetting on the new class split. On average, our Proxy-FDA consistently improves $\Delta_{\text{New}}$ for all prompt tuning methods, with competitive $\mathcal{A}_{\text{Base}}$ at the same time. Full results in Table~\ref{tb:base2new_full}.}
\vskip -0.1in
\label{tb:base2new_avg}
\tablestyle{+1pt}{1.1}
\addtolength{\tabcolsep}{+3pt}
\resizebox{0.8\linewidth}{!}{
\begin{tabular}{lc |cc |cc |cc |cc |cc |cc }
\toprule
 & \multicolumn{1}{c@{}}{} & \multicolumn{8}{c}{\textbf{Prompt tuning without regularization}} & \multicolumn{4}{c}{\textbf{Regularization-based}} \\ \cmidrule(lr){3-10} \cmidrule(lr){11-14}
 & \multicolumn{1}{c@{}}{} & \multicolumn{2}{c}{CoOp} & \multicolumn{2}{c}{CoCoOp} & 
\multicolumn{2}{c}{VPT} & \multicolumn{2}{c}{MaPLe} & \multicolumn{2}{c}{CLIPood} & \multicolumn{2}{c}{PromptSRC} \\
\cmidrule(lr){3-4} \cmidrule(lr){5-6} \cmidrule(lr){7-8} \cmidrule(lr){9-10} \cmidrule(lr){11-12} \cmidrule(lr){13-14}
~ & \multicolumn{1}{c@{}}{\textbf{+Proxy-FDA}} & \xmark & \multicolumn{1}{c@{}}{\cmark} & \xmark & \multicolumn{1}{c@{}}{\cmark} & \xmark & \multicolumn{1}{c@{}}{\cmark} & \xmark & \multicolumn{1}{c@{}}{\cmark} & \xmark & \multicolumn{1}{c@{}}{\cmark} & \xmark & \cmark \\
\midrule
\multirow{4}{*}{\shortstack[l]{Avg across\\ 11 datasets}} & $\mathcal{A}_{\text{Base}}$ & 82.69 &  \textbf{83.16}  &  \textbf{80.47}  &  80.36  &  \textbf{81.61}  &  81.55  &  82.28  &  \textbf{82.74}  &  83.91   & \textbf{84.33}  &  84.26  &  \textbf{84.47}  \\
& $\mathcal{A}_{\text{New}}$ & 63.22 & \textbf{73.67}  &  71.69  &  \textbf{76.44}  &  69.61  &  \textbf{73.89}  &  75.14   & \textbf{77.13}  &  74.50  &  \textbf{76.54}  &  76.10  &  \textbf{77.45}  \\
& $\Delta_{\text{New}}\uparrow$ & -10.99 &\textbf{-0.55} &-2.53 &\textbf{2.22}  &-4.61 &\textbf{-0.33} &0.92  &\textbf{2.91} & 0.28 & \textbf{2.33} & 1.88 & \textbf{3.23} \\
& $\mathcal{A}_{\text{H}}$ & 71.66  &  \textbf{78.13}  &  75.83  &  \textbf{78.35}  &  75.14  &  \textbf{77.53}  &  78.55  &  \textbf{79.84}  &  78.93  &  \textbf{80.25}  &  79.97  &  \textbf{80.81}  \\
\bottomrule
\end{tabular}
}
\end{table*}

\begin{figure*}[!t]
\begin{center}
\centerline{\includegraphics[width=1.0\linewidth]{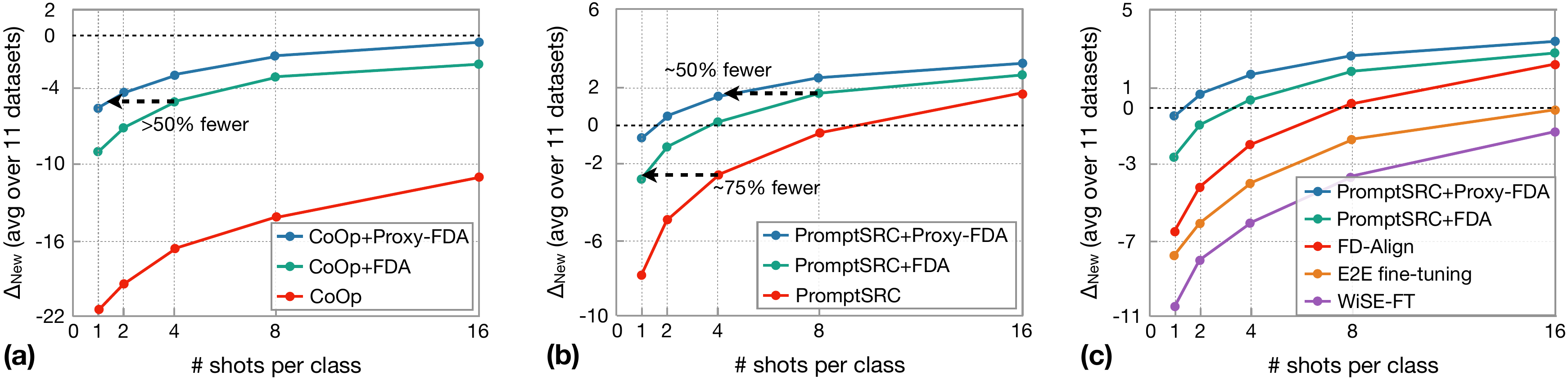}}
\end{center}
\vskip -0.3in
\caption{\textbf{(a-b)} The average $\Delta_{\text{New}}$ with varying number of shots per class for prompt tuning in the base-to-new setting. FDA achieves higher gains over the baselines in low-data regime, and our proxy learning further improves data efficiency. \textbf{(c)} PromptSRC+Proxy-FDA scales better with data than end-to-end fine-tuning and its improved variants (FD-Align and WiSE-FT) in the few-shot setting.}
\label{fig:few_shot_scaling}
\vskip -0.1in
\end{figure*}

\paragraph{Settings and metrics.} Prompt tuning is adopted for parameter-efficient fine-tuning in the few-shot scenario. We consider the two settings introduced in~\citep{zhou2022cocoop}: 1) \textbf{Base-to-new class generalization} within each dataset,~\ie,~prompt tuning on the base class split as $\mathcal{D}_{\text{ft}}$, and evaluating on the disjoint base and new class splits to obtain $\mathcal{A}_{\text{Base}}$ and $\mathcal{A}_{\text{New}}$. To quantify concept forgetting on the unseen new class split, we further report $\Delta_{\text{New}}$ as the change in $\mathcal{A}_{\text{New}}$ between pre-trained and prompt-tuned models -- the higher $\Delta_{\text{New}}$ the lower forgetting. 2) \textbf{Cross-dataset generalization} with ImageNet for prompt tuning and other 10 datasets for evaluation. Similarly, we report both the test accuracy $\mathcal{A}$ and accuracy change $\Delta_{\mathcal{A}}$ on each dataset to quantify forgetting. For all experiments, we report results as an average over three random seeds.

\paragraph{Implementation.} We apply our Proxy-FDA regularization to different prompt tuning baselines. For fair comparisons, we use the same implementation details of each baseline, including the prompt length, learning rate schedule and tuning epochs for each dataset. By default, all methods use 16 shots per class to prompt tune the CLIP model~\citep{radford2021learning} with ViT-B/16.

\paragraph{Results.} In Table~\ref{tb:base2new_avg}, we report results in the base-to-new setting. Proxy-FDA is applied to two categories of methods: 1) regularization-free prompt tuning baselines, which learn text prompts (CoOp~\citep{zhou2021coop}, CoCoOp~\citep{zhou2022cocoop}), image prompts (VPT~\citep{jia2022visual}) or both (MaPLe~\citep{khattak2023maple}). 2) regularization-based prompt learners. CLIPood~\citep{shu2023clipood} maintains a weighted ensemble of the pre-trained and fine-tuned models. State-of-the-art PromptSRC~\citep{khattak2023self} combines the ensembling strategy with both feature- and logit- level regularization between the original and fine-tuned models (but in a point-wise manner).

We can see from Table~\ref{tb:base2new_avg} that, averaged across 11 datasets, Proxy-FDA consistently improves the $\mathcal{A}_{\text{New}}$ of all regularization-free baselines, sometimes by a large margin (10.45 for CoOp), with competitive $\mathcal{A}_{\text{Base}}$ at the same time. The gains in $\mathcal{A}_{\text{New}}$ translate to gains in $\Delta_{\text{New}}$, indicating the utility of Proxy-FDA in lowering forgetting for few-shot settings. The per-dataset results in Table~\ref{tb:base2new_full} (in Appendix) show that $\Delta_{\text{New}}$ sees particularly large gains on 3 semantically distant datasets (DTD, EuroSAT and UCF101), thanks to our strong capability of preserving pre-trained knowledge. Overall, Proxy-FDA boosts the $\mathcal{A}_{\text{H}}$ of MaPLe to 79.84, being already better than or on par with that of the regularization methods CLIPood (78.93) and PromptSRC (79.97). Encouragingly, Proxy-FDA is complementary to the two regularization methods and can further improve them in all metrics.

Fig.~\ref{fig:few_shot_scaling} shows the superior data efficiency of Proxy-FDA when lowering forgetting for few-shot prompt tuning. In the base-to-new setting, we vary the amount of tuning data and find that the $\Delta_{\text{New}}$ gain of FDA increases with less data. Meanwhile, our proxy learning component further improves data efficiency, often matching the FDA performance on half the data. Fig.~\ref{fig:few_shot_scaling}(c) also shows the benefits of (Proxy-)FDA over end-to-end fine-tuning and its improved variants --- FD-Align~\citep{song2023fdalign} and WiSE-FT~\citep{wortsman2022robust} --- in data-limited regimes.

In the Appendix, Table~\ref{tb:cross_dataset} further shows results under the cross-dataset generalization setting. Proxy-FDA is shown to prevent concept forgetting consistently, with uniformly increased $\Delta_{\mathcal{A}}$ on target datasets and a good trade-off with $\mathcal{A}$ on the source dataset ImageNet. Table~\ref{tb:Prompt_learning_2024} shows our advantage over more recent prompt tuning methods that benefit from either LLM or advanced regularization techniques.

\begin{table*}[!t]
\caption{\textbf{Continual fine-tuning: test accuracy $\mathcal{A}_{\text{LP}}$ and $\Delta_{\text{LP}}$ for models fine-tuned on three task sequences}.
The first 3 rows show performance on fine-tuned tasks and the 4th row shows performance averaged on 6 other datasets.}
\label{tb:continual_fine_tuning}
\begin{center}
\vskip -0.2in
\resizebox{1.0\linewidth}{!}{
\begin{tabular}{ll|cc|cc|cc|cc|cc|cc}
\toprule
Fine-tune & \multicolumn{1}{l@{}}{Evaluation} & \multicolumn{2}{c}{Naive End-to-End} & \multicolumn{2}{c}{LP-FT} & \multicolumn{2}{c}{L2SP} & \multicolumn{2}{c}{LDIFS} & \multicolumn{2}{c}{FDA (ours)} & \multicolumn{2}{c}{Proxy-FDA (ours)} \\
\cmidrule(lr){3-4} \cmidrule(lr){5-6} \cmidrule(lr){7-8} \cmidrule(lr){9-10} \cmidrule(lr){11-12} \cmidrule(lr){13-14}
dataset& \multicolumn{1}{l@{}}{dataset} & $\mathcal{A}_{\text{LP}}$ & \multicolumn{1}{c@{}}{$\Delta_{\text{LP}}\uparrow$}& $\mathcal{A}_{\text{LP}}$ & \multicolumn{1}{c@{}}{$\Delta_{\text{LP}}\uparrow$} & $\mathcal{A}_{\text{LP}}$ & \multicolumn{1}{c@{}}{$\Delta_{\text{LP}}\uparrow$} & $\mathcal{A}_{\text{LP}}$ & \multicolumn{1}{c@{}}{$\Delta_{\text{LP}}\uparrow$} & $\mathcal{A}_{\text{LP}}$ & \multicolumn{1}{c@{}}{$\Delta_{\text{LP}}\uparrow$} & $\mathcal{A}_{\text{LP}}$ & $\Delta_{\text{LP}} \uparrow$\\
\midrule
\multirow{4}{*}{\shortstack[l]{SVHN$\rightarrow$\\CIFAR10$\rightarrow$\\RESISC45}} & SVHN & 90.29 &-7.13 &90.97 &-6.46 & 91.93 &-4.53 &96.68& -0.41 &\textbf{96.77} &0.61 &96.72 &\textbf{0.93}\\
& CIFAR10  & 95.25 &-2.31 &96.31 &-1.57 &97.26 &-0.25 &\textbf{97.41}& -0.21 &97.13 &0.57 &97.29 &\textbf{1.02}\\
& RESISC45 & 95.30 &4.00 &94.29 &2.98 &93.44 &2.16 &95.00 &3.70 &95.22 &4.14 &\textbf{95.38} &\textbf{4.22}\\ \cmidrule(lr){2-14}
& Others & 80.91 &-5.08 &82.13 &-4.24 &86.89 &-0.01 &87.08 &0.10 &\textbf{87.21} &0.76 &86.95 &\textbf{1.08} \\ \midrule
\multirow{4}{*}{\shortstack[l]{SVHN$\rightarrow$\\CIFAR100$\rightarrow$\\RESISC45}} & SVHN & 90.05 &-7.28 &94.42 &-2.73 & 90.42 &-6.12 &96.32& -0.65 &96.18 &0.63 &\textbf{96.43} &\textbf{0.71}\\
& CIFAR100  & 81.08 &-7.18 &82.63 &-3.04 &85.72 &-0.88 &\textbf{86.54}& -0.30 &86.33 &0.72 &86.14 &\textbf{0.85}\\
& RESISC45 & 95.40 &\textbf{4.13} &93.81 &2.51 &93.21 &1.90 &95.11 &3.83 &95.32 &3.95 &\textbf{95.46} &4.01\\ \cmidrule(lr){2-14}
& Others & 83.76 &-4.65 &85.14 &-4.02 &89.04 &-0.37 &\textbf{89.12} &-0.23 &89.02 &0.68 &89.09 &\textbf{0.96} \\ \midrule
\multirow{4}{*}{\shortstack[l]{SVHN$\rightarrow$\\Cars$\rightarrow$\\RESISC45}} & SVHN & 95.93 &-1.45 &96.58 &-0.76 & 95.98 &-0.44 &96.90& -0.17 &96.74 &0.79 &\textbf{96.91} &\textbf{0.94}\\
& Cars  & 76.96 &-4.18 &71.60 &-8.36 &81.82 &-0.40 &84.23& 0.47 &\textbf{84.38} &1.14 &84.32 &\textbf{1.36}\\
& RESISC45 & 95.17 &3.89 &94.35 &3.00 &93.43 &2.13 &\textbf{95.27} &3.73 &95.12 &3.92 &95.23 &\textbf{4.07}\\ \cmidrule(lr){2-14}
& Others & 83.38 &-4.93 &84.39 &-4.51 &87.15 &-1.67 &89.39 &0.23 &89.54 &0.96 &\textbf{89.67} &\textbf{1.17} \\
\bottomrule
\end{tabular}
}
\end{center}
\vskip -0.1in
\end{table*}

\subsection{Continual Fine-tuning}

Finally, we perform continual fine-tuning and see whether we can learn a sequence of downstream tasks without forgetting concepts. We follow~\citep{mukhoti2024finetuning} to train on three task sequences: SVHN$\rightarrow$CIFAR10$\rightarrow$RESISC45, SVHN$\rightarrow$CIFAR100$\rightarrow$RESISC45 and SVHN$\rightarrow$Cars$\rightarrow$ RESISC45. Table~\ref{tb:continual_fine_tuning} shows our FDA and Proxy-FDA methods progressively improve the $\Delta_{\text{LP}}$ for each task sequence, both achieving positive forward transfer with all positive $\Delta_{\text{LP}}$ values.
Proxy-FDA always attains the highest $\Delta_{\text{LP}}$ (except on RESISC45 in the second sequence), while still remaining competitive in $\mathcal{A}_{\text{LP}}$.
Table~\ref{tb:continual_baseline} and~\ref{tb:continual_SplitImageNetR} in Appendix~\ref{sec:appendix_more_results} show our benefits over popular continual learning baselines for both the 3-task setup and the class-incremental setting on Split ImageNet-R~\citep{wang2022dualprompt}.

\subsection{Applications Beyond Classification}

Appendix~\ref{sec:appendix_more_tasks} shows that the benefits of Proxy-FDA hold for fine-tuning tasks beyond classification. We consider the vision-language tasks of image captioning and VQA, where Proxy-FDA outperforms baselines in mitigating forgetting. We further show Proxy-FDA is applicable to knowledge distillation and achieves quite promising performance.

\section{Conclusion}

This paper introduces Proxy-FDA, a novel feature-space regularization method that preserves concepts during fine-tuning. The core idea is to align the local structures of pre-trained and fine-tuned feature distributions with learned proxies. A structure-aware distributional distance metric is used to assess the feature alignment quality, demonstrating a strong correlation with concept forgetting. Our approach achieves state-of-the-art results in mitigating forgetting in various fine-tuning settings and across different tasks.


\section*{Impact Statement}
The main contribution of this work is a new feature-space regularization method for robust fine-tuning. Our method is shown to effectively preserve the concepts in pre-trained vision foundation models. One potential societal impact is that, when the pre-trained concepts reflect (unintentional) biases, our regularization method could inherit or amplify those biases in fine-tuned features. As a result, one may observe perpetual unfair or discriminative outcomes in downstream tasks and more critical applications such as AI-driven planning and decision-making.

\bibliography{myref}
\bibliographystyle{icml2025}

\newpage
\appendix
\onecolumn

\begin{figure}[!t]
\begin{center}
\centerline{\includegraphics[width=0.83\linewidth]{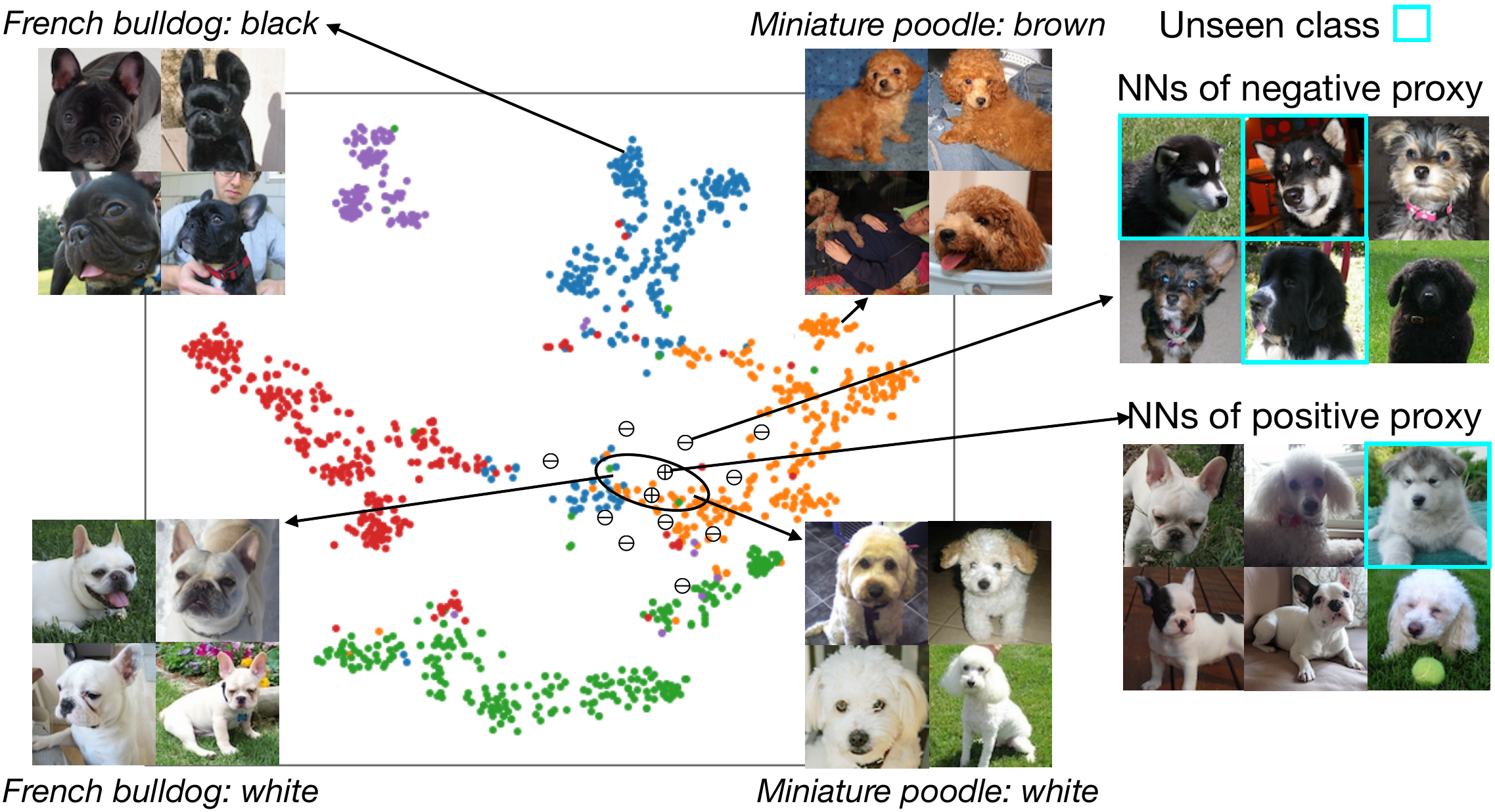}}
\end{center}
\vskip -0.3in
\caption{t-SNE visualization of the local feature neighborhood (circled) on ImageNet for the pre-trained CLIP ViT-B/16 model. In this neighborhood, we observe the same white color from two dog breeds ``French bulldog'' and ``Miniature poodle''. Preserving CLIP's common-sense knowledge (in this case the color attribute shared across different classes) using FDA maintains the generalizability of foundation models. On the other hand, the generated proxies include diverse information from both seen and unseen (\eg,~``Malamute'') classes that can regularize the neighborhood boundary and further improve FDA. The synthesized seen/unseen class data are illustrated by kNN retrieval from the base/new class splits of ImageNet when fine-tuning on the base only.} 
\label{fig:tsne}
\end{figure}

\begin{figure}[!t]
\begin{center}
\centerline{\includegraphics[width=0.5\linewidth]{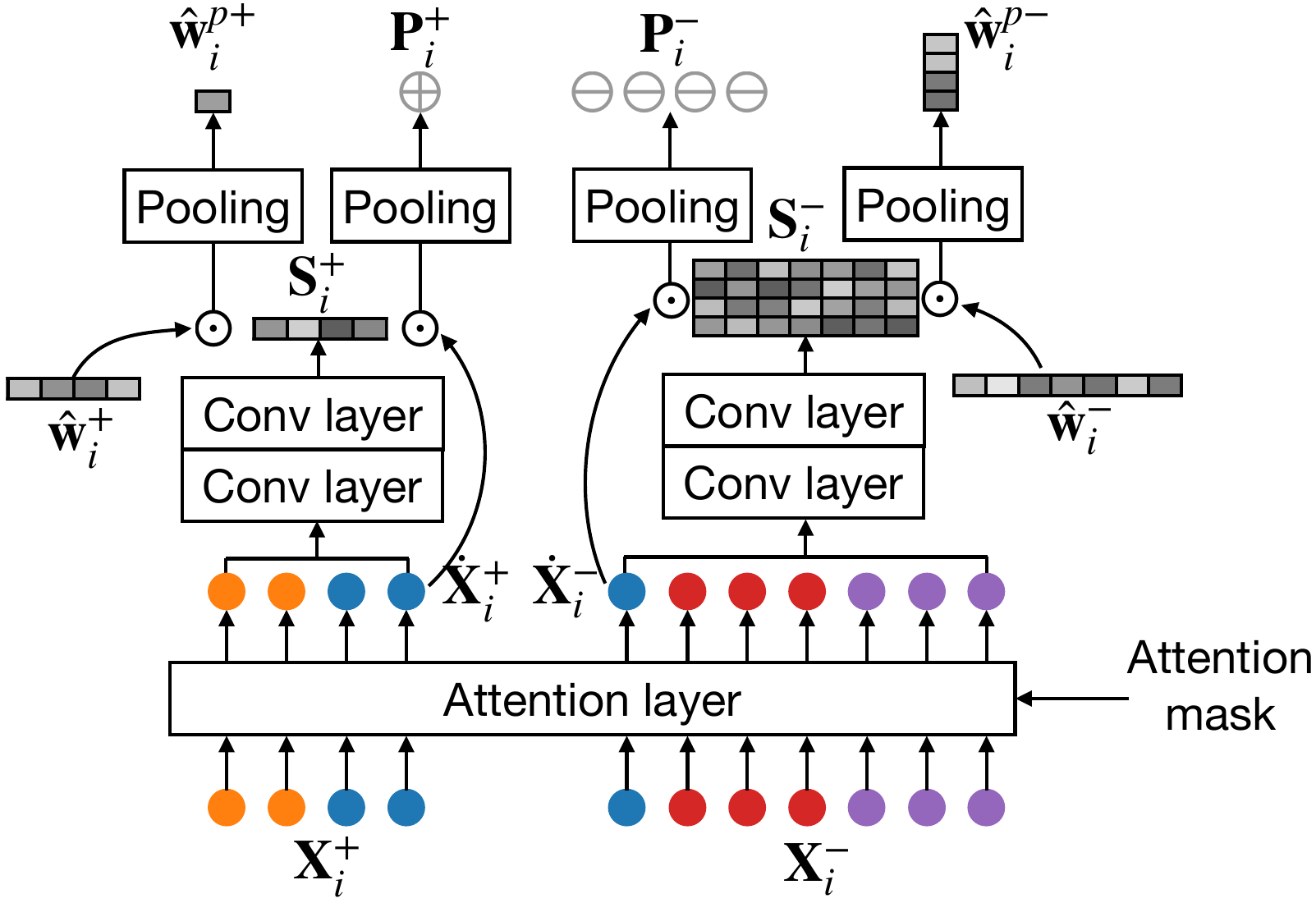}}
\end{center}
\vskip -0.35in
\caption{Efficient architecture of our proxy generator that generates dynamic proxies or synthetic features.} 
\label{fig:proxy_generator}
\vskip -0.1in
\end{figure}

\section{Hard Class Mining}
\label{sec:hard_mining}

As mentioned in the main text (Section~\ref{sec:fda_baseline}), we perform hard class mining in the mini-batch to facilitate the modeling and alignment of local neighborhood structures. The high-level idea of hard class mining is to greedily select class distributions that are close to one another. More specifically, we construct our mini-batch in the following way:
\vspace{-0.1in}
\begin{enumerate}[leftmargin=20pt]
\setlength{\itemsep}{0pt}
\item Randomly choose a large number of classes $C \gg m$; for each class, randomly sample $n$ examples to extract their feature embeddings using both $f_{\hat\theta}$ and $f_{\theta}$.
\item Sample a seed class randomly from the $C$ classes. Then greedily add a new class that has the largest class-wise loss $\sum_{i=1}^n \mathcal{L}_{\text{FDA}}^i$ (Eq.~(2)) \emph{w.r.t.} the selected classes till we reach $m$ classes. Note in this greedy process, we set the neighborhood size $K=n$ when computing $\mathcal{L}_{\text{FDA}}^i$.
\item Construct batch with the selected $m$ classes, each with $n$ examples.
\end{enumerate}
\vspace{-0.1in}

\section{Efficient Architecture of Instance-wise Proxy Generator}
\label{sec:proxy_generator}

Fig.~\ref{fig:proxy_generator} shows the network architecture of our proxy generator that is trained online using Eq.~(3-4). The input $\mX_{i}^{+}$ and $\mX_{i}^{-}$ first go through an attention layer to model the global context within each set and fuse features thoroughly. Attention mask is used to ensure the independence between the two sets. Next, we dynamically pool the intermediate features $\dot{\mX}_{i}^{+} \in \mathbb{R}^{d \times K}$ and $\dot{\mX}_{i}^{-} \in \mathbb{R}^{d \times (B-K-1)}$ via learned pooling functions, as summarized below. Through such pooling, we can predict proxies $\{\mP_{i}^{+},\mP_{i}^{-}\}$ and their similarity estimates $\{\hat\vw_{i}^{p+},\hat\vw_{i}^{p-}\}$ all at once. 
\begin{eqnarray}
\textbf{Predict pooling weights:}\!\!\!\!\!\!\!\! &&\mS_i^+=h^+(\dot{\mX}_{i}^{+})\in \mathbb{R}^{K \times n^{p+}}, \;\;\;\mS_i^-=h^-(\dot{\mX}_{i}^{-})\in \mathbb{R}^{(B-K-1) \times n^{p-}},\\
\textbf{Pooling in matrix form:}\!\!\!\!\!\!\!\! &&\mP_i^+=\dot{\mX}_{i}^{+} \cdot \mS_i^+ \in \mathbb{R}^{d \times n^{p+}}, \;\;\; \mP_i^-=\dot{\mX}_{i}^{-} \cdot \mS_i^- \in \mathbb{R}^{d \times n^{p-}}, \\
\!\!\!\!\!\!\!\! &&\hat\vw_{i}^{p+}= {\mS_i^+}^T \cdot \hat\vw_{i}^{+}\in \mathbb{R}^{n^{p+}}, \;\hat\vw_{i}^{p-}= {\mS_i^-}^T \cdot \hat\vw_{i}^{-}\in \mathbb{R}^{n^{p-}}, \\
\text{where}\!\!\!\!\!\!\!\! &&\;\;\;\;\;\;\;\;\;\;\;\;\;\;\;\;\;\;\hat\vw_{i}^{+}\in \mathbb{R}^{K}, \;\;\;\;\;\;\;\;\;\;\;\;\;\;\;\;\;\;\;\;\;\;\;\;\hat\vw_{i}^{-}\in \mathbb{R}^{B-K-1}. \notag
\label{eq_pooling}
\end{eqnarray}

Note both $h^+(\cdot)$ and $h^-(\cdot)$ are implemented by two convolutional layers, but with different output channel sizes ($n^{p+}$ and $n^{p-}$ respectively). The output pooling weights $\mS_i^+$ and $\mS_i^-$ are softmax-normalized, leading to convex combinations of features and feature similarities during the pooling stage. This eases training of pooling functions and makes sure the pooled results are valid (especially the pooled similarity estimates).

\section{Distributional Distance Metric: OTDD}
\label{sec:otdd}

To measure FDA quality, there are many distance metrics for distribution alignment. Here we choose the distributional distance metric based on Optimal Transport Dataset Distance (OTDD)~\citep{NEURIPS2020_f52a7b26}. OTDD is especially suited to measure the alignment quality of feature distributions with local structures, because this distance metric takes both the label distribution and clustering structure of the feature distributions into consideration.

Specifically, OTDD uses the feature and label distributions $(\vx,y)|_{\vx\in \mathcal{X}, y\in \mathcal{Y}}$ to compute the distance between two datasets. Given that the source and target datasets may have different label sets, the high-level idea of OTDD is to represent each class label as a distribution over the in-class features. This transforms the source and target label sets into the shared space of feature distributions over $\mathcal{X}$. In our context of model fine-tuning, we have pre-trained features $\hat\vx$ and fine-tuned features $\vx$ that are likely shifted from $\hat\vx$. They form the source and target feature distributions respectively, and have different labels $\hat y$ and $y$ (details later). Then we can define the label distance $D_{\mathcal{Y}}(\hat y,y)$ using the $p$-Wasserstein distance associated with the L2 distance $\|\hat\vx-\vx\|_{2}^{2}$ in $\mathcal{X}$. This enables one to measure the distributional difference in $\mathcal{X} \times \mathcal{Y}$:
\begin{equation}
    D_{\mathcal{X} \times \mathcal{Y}}\left( (\hat\vx,\hat y), (\vx,y)\right) = \left( D_{\mathcal{X}}(\hat\vx-\vx)^p + D_{\mathcal{Y}}(\hat y,y)^p \right)^{1/p}.
\label{eq:otdd}
\end{equation}
Please refer to~\citep{NEURIPS2020_f52a7b26} for the exact formulation. To capture the clustering structure of both the pre-trained and fine-tuned feature distributions, we perform K-Means clustering per class on each feature distribution. This results in pseudolabels $\hat y$ and $y$ that are more fine-grained than class labels for OTDD computation.

\begin{figure}[!t]
\begin{center}
\centerline{\includegraphics[width=0.95\linewidth]{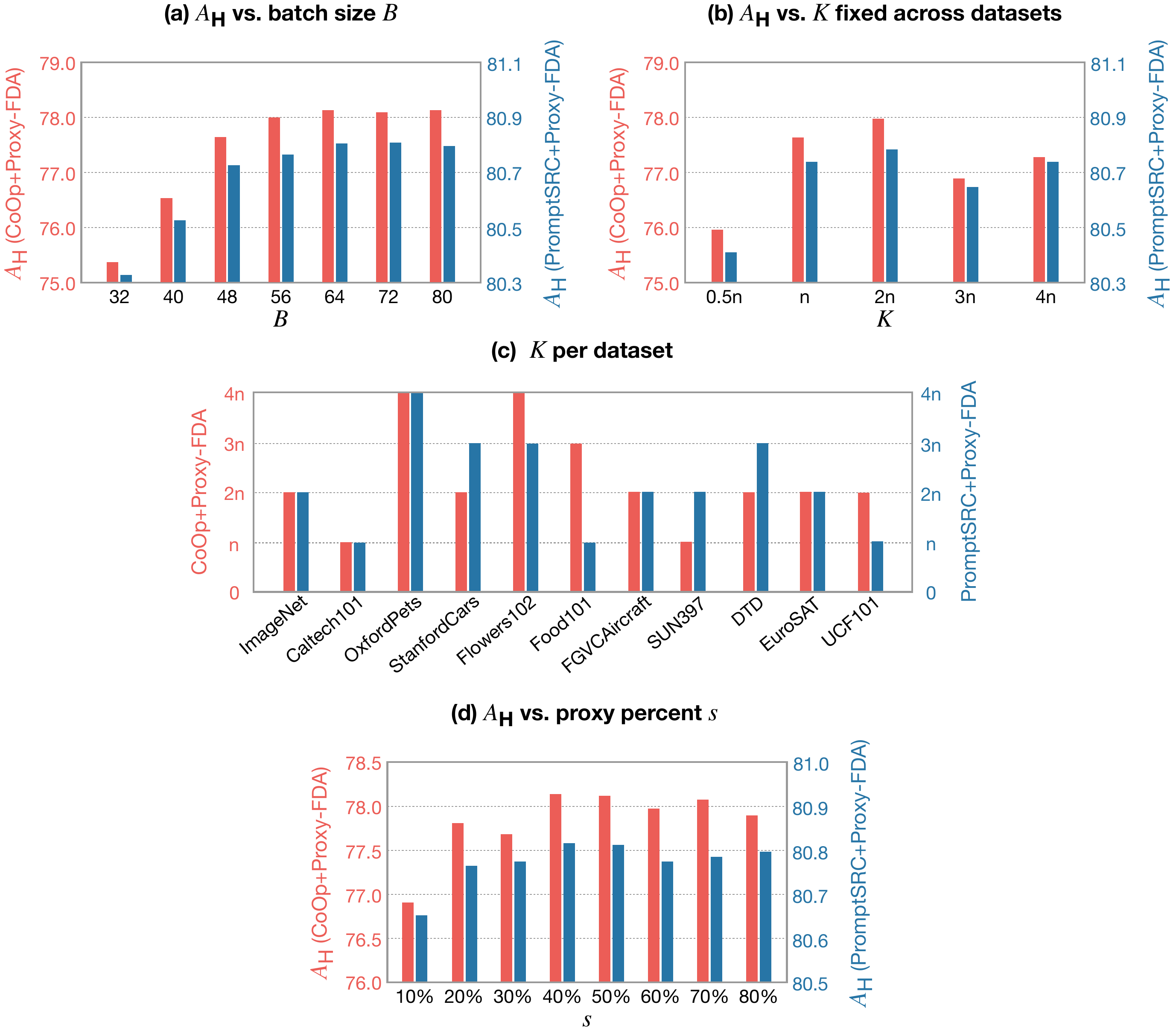}}
\end{center}
\vskip -0.3in
\caption{\textbf{Sensitivity analysis for hyper-parameters:} \textbf{(a)} batch size $B$, \textbf{(b)} neighborhood size $K$ that is fixed across datasets, \textbf{(c)} optimal $K$ per dataset, and \textbf{(d)} scalar $s$ that decides the percent number of generated proxies compared to that of real samples.
Analysis is performed for few-shot prompt tuning in the base-to-new setting (16 shots per class). We report the $\mathcal{A}_{\text{H}}$ averaged across 11 datasets, when applying Proxy-FDA to two representative baselines CoOp and PromptSRC.
Note $\mathcal{A}_{\text{H}}$ is the Harmonic mean of $\mathcal{A}_{\text{Base}}$ (representing prompt-tuning accuracy itself) and $\mathcal{A}_{\text{New}}$ (representing generalization and can derive $\Delta_{\text{New}}$). Hence $\mathcal{A}_{\text{H}}$ is ideal for hyper-parameter sweeping since $\mathcal{A}_{\text{H}}$ denotes a trade-off between downstream accuracy and concept forgetting ($\Delta_{\text{New}}$).}
\label{fig:hyperparameters}
\vskip -0.1in
\end{figure}

\section{Analysis of Hyper-parameters}
\label{sec:hyper-parameters}

\paragraph{Hyper-parameters.} Fig.~\ref{fig:hyperparameters}(a) shows our Proxy-FDA approach benefits from a relatively large batch size $B$ to preserve meaningful structures of feature neighborhoods. Performance decreases when $B<64$; when $B$ grows larger than 64, performance seems quite robust to varying batch size. By default, we set $B=64$ that best fits in our GPU memory.

Based on the hard class mining strategy (Section~\ref{sec:hard_mining}), we construct a mini-batch with $m=16$ hard-mined classes, each with $n=4$ class samples. Note in few-shot settings, each class may not have enough data ($<4$) for sampling,~\eg,~only 1 or 2 shots are available per class. In this case, we perform random data augmentation to guarantee $n=4$ samples per class.
On the other hand, a relatively large $m$ ensures diverse class distributions in a batch, which allows better characterization of local feature neighborhoods. Diverse classes also allow pooling rich proxies from them, resulting in unseen data variations or new class concepts to further improve FDA.

Our Proxy-FDA method has two key hyper-parameters: the neighborhood size $K>n$ and a scalar $s$. The latter makes the number of positive proxies $n^{p+}=s\cdot K$ and negative proxies $n^{p-}=s\cdot (B-K-1)$ proportional to the set size of the true positives $\mX_{i}^{+} \in \mathbb{R}^{d \times K}$ and true negatives $\mX_{i}^{-} \in \mathbb{R}^{d \times (B-K-1)}$.

The intuition of setting $K>n$ is to identify sufficient neighbors from more than one class, for meaningful FDA between similar clusters of related classes. Nevertheless, the exact value of $K$ is varied as a function of dataset distribution, as each dataset has different levels of intra- and inter-class variation. In practice, we pick the best $K$ per dataset from $\{n,2n,3n,4n\}$. Fig.~\ref{fig:hyperparameters}(b) shows how performance generally varies with $K$ when $K$ is fixed across 11 datasets. We see that $K=2n$ works best, while it noticeably hurts performance when $K<n$, confirming our intuition above. Hence we stick to the constraint of $K>n$ for per-dataset $K$ selection (Fig.~\ref{fig:hyperparameters}(c)).

On the other hand, the scalar $s$ is set to 0.4 by default. This leads to a virtual batch size of around 90 (increased from 64). The virtual batch now consists of true and synthetic features for FDA. Fig.~\ref{fig:hyperparameters}(d) shows the sensitivity analysis for $s$.

Lastly, the weighting parameter for $\mathcal{L}_{\text{var}}$ (Eq.~(3-4)) is fixed at $\alpha=5$ for all experiments. We observe no meaningful improvements via more careful tuning of $\alpha$. The weighting parameter $\lambda$ is used to balance the task loss against our regularization loss (Eq.~(\ref{eq1}) and (\ref{eq:proxyfda})). We tune $\lambda$ on a held-out validation set of each dataset.


\begin{figure}[!t]
\begin{center}
\centerline{\includegraphics[width=1.0\linewidth]{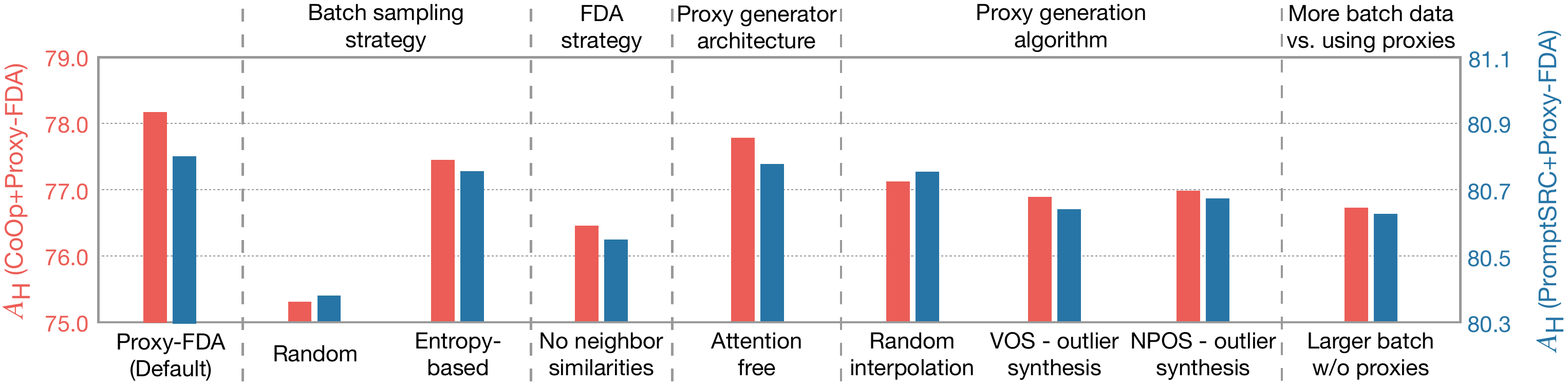}}
\end{center}
\vskip -0.2in
\caption{\textbf{Ablating the Proxy-FDA components} based on few-shot prompt tuning in the base-to-new setting (16 shots per class). We report $\mathcal{A}_{\text{H}}$ averaged across 11 datasets, when applying Proxy-FDA to two representative baselines CoOp and PromptSRC.
Note $\mathcal{A}_{\text{H}}$ is the Harmonic mean of $\mathcal{A}_{\text{Base}}$ (representing prompt-tuning accuracy itself) and $\mathcal{A}_{\text{New}}$ (representing generalization and can derive $\Delta_{\text{New}}$). Hence $\mathcal{A}_{\text{H}}$ is ideal for ablation studies since $\mathcal{A}_{\text{H}}$ denotes a trade-off between downstream accuracy and concept forgetting ($\Delta_{\text{New}}$).
For the proxy generation strategy, we compare with random linear interpolation~\citep{verma19a} and outlier feature synthesis methods VOS~\citep{du2022vos} and NPOS~\citep{tao2023nonparametric}.}
\label{fig:Ablations}
\end{figure}

\begin{table*}[!t]
\vskip 0.2in
\caption{\textbf{Quantifying proxy diversity} using the variance loss in Eq.~(3-4). Particularly, we report the diversity metric as the average standard deviation term $1/d \sum_{j=1}^{d} \sqrt{\text{Var}(\mP_{j,:})}$ in the variance loss: higher value indicates larger proxy diversity. To further aggregate the metrics of the positive and negative proxies $\{\mP_{i}^{+},\mP_{i}^{-}\}$, we take their mean and compute its moving average till fine-tuning is completed. We compare the aggregated diversity metric of all the proxy generation methods as ablated in Fig.~\ref{fig:Ablations}.}
\label{tb:proxy_diversity}
\begin{center}
\vskip -0.1in
\resizebox{0.8\linewidth}{!}{
\begin{tabular}{ccccc}
\toprule
& Proxy generation (ours) & Random interpolation & VOS & NPOS \\
\midrule
Diversity metric $\times 10^{-2}$ & \textbf{3.14} & 2.89 & 1.53 & 1.72 \\
\bottomrule
\end{tabular}
}
\end{center}
\end{table*}

\section{Ablating Proxy-FDA Components}
\label{sec:ablating_proxyfda}

Fig.~\ref{fig:Ablations} includes ablation studies on the key components of Proxy-FDA, in the few-shot prompt tuning setting.

\paragraph{Batch sampling strategy.} We start with comparing the default hard class mining method with random class sampling. Their considerable performance difference shows that hard class mining is crucial. Indeed, one can better model the nearest neighbor graphs from close class samples, which facilitates the following graph matching for FDA purpose. We further compare with an entropy-based batch sampling strategy that prioritizes similar class samples simply by low entropy. This sampling strategy is found less performant, likely because entropy cannot characterize sample similarity adaptively as a function of current \textit{feature distribution structure}. As a result, such batch sampling criterion is decoupled with the structural FDA within sampled batch. While our default strategy samples similar classes directly using FDA loss, which could adapt to the feature distribution structure, and is coupled with FDA in batch.

\paragraph{FDA strategy.} One may wonder what if we only align the neighbor indices during FDA, without considering the neighbor similarities (\ie,~keeping $\hat w_{ij}=1$)? We see that this baseline leads to large performance drop, demonstrating that both neighbor indices and similarities are indispensable for effective FDA.

\paragraph{Proxy generator architecture.} We first note that our proxy generator is learned to produce unseen data out of diverse feature combinations within the positive set $\mX_{i}^{+}$ or negative set $\mX_{i}^{-}$. The attention layer helps to achieve this goal by modeling the global context among all input features with pairwise attention. Convolutional layers, however, only have local receptive fields and have to rely on pooling operations to capture long-range dependencies. Here we compare with an attention-free architecture that has the attention layer replaced with convolutional plus pooling layers -- the resulting proxy generator maintains a similar parameter count. The attention-free architecture is observed to achieve consistently lower performance, likely due to the lower quality of generated proxies.

\paragraph{Proxy generation algorithm.} We compare with three baselines. One simple method is based on linear interpolation between random feature pairs from both $\mX_{i}^{+}$ and $\mX_{i}^{-}$. Feature similarity estimates are interpolated in the same way. We see random interpolation obtains inferior performance than our learning-based approach. This is because our approach can learn to synthesize informative proxies that best help FDA: the diverse proxies can not only enrich data but also refine the decision boundary between positive and negative feature manifolds. This is not possible with random interpolation. On the other hand, the parametric VOS and non-parametric NPOS methods learn to synthesize outlier features only in low-likelihood regions (often around decision boundaries between classes). The two methods are observed to achieve even worse results than random interpolation. We conjecture that this is because outliers are not able to encode diverse unseen data/concepts that are crucial for improving FDA.

Lastly, we quantify the proxy feature diversity in Table~\ref{tb:proxy_diversity} using our variance loss. Interestingly, it is observed that the diversity metric of a proxy generation method highly correlates with its performance: our proxy generation method outperforms random interpolation in both proxy diversity and final accuracy. This trend also holds when comparing our method with VOS/NPOS.

\paragraph{Using more batch data or proxies.} To further quantify the effect of proxy learning that virtually increases the batch size $B$ from 64 to around 90, we compare with FDA simply on a larger batch with a similar number of true feature points. Specifically, we construct the batch with $m=22$ hard-mined classes, each with $n=4$ examples. Hence the batch size is comparable to that of Proxy-FDA, but without proxies. We observe from Fig.~\ref{fig:Ablations} that simply using a larger batch size does not perform as well. Instead, it is worth using our proxy generator to increase data diversity with only a small overhead.

\section{More Results}
\label{sec:appendix_more_results}

\begin{table*}[!t]
\caption{\textbf{Test accuracy $\mathcal{A}_{\text{LP}}$ of end-to-end fine-tuned model on ImageNet and its average $\Delta_{\text{LP}}$ computed over 5 datasets (DTD, EuroSAT, GTSRB, RESISC45 and SVHN)}.
We study different architectures of CLIP~\citep{radford2021learning}, FLAVA~\citep{singh2022flava}, DINOv2~\citep{oquab2024dinov} and MAE~\citep{He2022}.
$\Delta_{\text{LP}}$ denotes the change in $\mathcal{A}_{\text{LP}}$ between pre-trained and fine-tuned models on target dataset, quantifying the level of concept forgetting. Higher $\Delta_{\text{LP}}$ shows lower forgetting or even positive forward transfer ($\Delta_{\text{LP}}>0$).
Note we initialize the model's linear head with zero-shot weights for naive fine-tuning, and with Linear Probe (LP) weights for all other methods including ours. The initialized zero-shot weights are the text encodings of class name for CLIP and FLAVA, and random weights for DINOv2 and MAE.}
\vskip -0.1in
\label{tb:e2e_across_architectures}
\begin{center}
\resizebox{0.9\linewidth}{!}{
\begin{tabular}{ll|cc|cc|cc|cc|cc|cc}
\toprule
Model & \multicolumn{1}{l@{}}{Architecture} & \multicolumn{2}{c}{Naive End-to-End} & \multicolumn{2}{c}{LP-FT} & \multicolumn{2}{c}{L2SP} & \multicolumn{2}{c}{LDIFS} & \multicolumn{2}{c}{FDA (ours)} & \multicolumn{2}{c}{Proxy-FDA (ours)} \\
\cmidrule(lr){3-4} \cmidrule(lr){5-6} \cmidrule(lr){7-8} \cmidrule(lr){9-10} \cmidrule(lr){11-12} \cmidrule(lr){13-14}
& \multicolumn{1}{c@{}}{} & $\mathcal{A}_{\text{LP}}$ & \multicolumn{1}{c@{}}{$\Delta_{\text{LP}}\uparrow$}& $\mathcal{A}_{\text{LP}}$ & \multicolumn{1}{c@{}}{$\Delta_{\text{LP}}\uparrow$} & $\mathcal{A}_{\text{LP}}$ & \multicolumn{1}{c@{}}{$\Delta_{\text{LP}}\uparrow$} & $\mathcal{A}_{\text{LP}}$ & \multicolumn{1}{c@{}}{$\Delta_{\text{LP}}\uparrow$} & $\mathcal{A}_{\text{LP}}$ & \multicolumn{1}{c@{}}{$\Delta_{\text{LP}}\uparrow$} & $\mathcal{A}_{\text{LP}}$ & $\Delta_{\text{LP}} \uparrow$\\
\midrule
\multirow{4}{*}{CLIP} & ResNet-50 & 78.39 &-4.01 &78.45 &-3.40 & 76.13 &-1.54 &78.16& -0.11 &78.43 &0.62 &\textbf{78.58} &\textbf{0.89}\\
& ViT-B/32  & 82.02 &-3.02 &82.12 &-2.17 &80.78 &-0.88 &\textbf{82.21}& 0.10 &81.93 &0.81 &82.16 &\textbf{1.15}\\
& ViT-B/16 & 85.21 &-2.92 &85.36 &-1.73 &82.19 &-0.74 &85.31 &0.16 &\textbf{85.41} &0.92 &85.40 &\textbf{1.03}\\
& ViT-L/14 & 87.88 &-2.33 &87.91 &-1.52 &86.87 &-0.43 &87.85 &0.22 &\textbf{87.99} &1.02 &87.96 &\textbf{1.28} \\ \midrule
FLAVA & ViT-B/16 & 81.18 &-3.94 &81.36 &-3.04 &80.11 &-1.10 &\textbf{81.61} &0.04 &81.47 &0.61 &81.59 &\textbf{0.96} \\ \midrule
\multirow{2}{*}{DINOv2} & ViT-B/14 & 85.32 &-2.71 &85.48 &-1.86 &84.50 &-0.66 &86.02 &0.06 &86.23 &0.68 &\textbf{86.34} &\textbf{0.85} \\
& ViT-L/14 & 87.60 &-1.92 &87.90 &-1.40 &87.02 &-0.19 &\textbf{87.91} &0.13 &87.87 &0.77 &87.71 &\textbf{0.94} \\ \midrule
\multirow{2}{*}{MAE} & ViT-B/16 & 83.57 &-5.10 &83.81 &-4.36 &82.84 &-3.03 &83.76 &-0.94 &83.73 &-0.08 &\textbf{83.94} &\textbf{0.39} \\
& ViT-L/16 & 85.86 &-4.26 &\textbf{86.04} &-3.59 &85.10 &-1.82 &85.90 &-0.12 &85.86 &0.79 &85.67 &\textbf{0.94}\\
\bottomrule
\end{tabular}
}
\end{center}
\end{table*}

\paragraph{End-to-end fine-tuning.} Table~\ref{tb:e2e_across_architectures} shows ImageNet fine-tuning results with different foundation models and architectures. We see that both FDA and Proxy-FDA consistently improve the $\Delta_{\text{LP}}$ over other baselines, with Proxy-FDA offering the highest $\Delta_{\text{LP}}$ values. This comes with competitive downstream accuracy $\mathcal{A}_{\text{LP}}$ on ImageNet. Notably, our obtained $\Delta_{\text{LP}}$ values are mostly positive, with a sole exception of MAE model (ViT-B/16 architecture) when fine-tuned using FDA. This indicates that we can achieve positive forward transfer in most cases and otherwise minimized concept forgetting.

\paragraph{Few-shot prompt-tuning.} Table~\ref{tb:base2new_full} lists the full results of prompt tuning on each of the 11 datasets under the base-to-new class generalization setting. Table~\ref{tb:cross_dataset} shows results under the cross-dataset generalization setting,~\ie,~quantifying generalization from ImageNet to 10 target datasets. In both settings, Proxy-FDA is plugged into different prompt tuning baselines. Proxy-FDA is observed to reduce concept forgetting consistently on unseen data with comparable performance on seen data.

We further compare with more recent prompt tuning methods in Table~\ref{tb:Prompt_learning_2024}. Comparisons are conducted under the base-to-new class generalization setting, and an additional domain generalization setting. In the latter setting, we prompt tune on ImageNet (16 shots per class) and evaluate OOD generalization on ImageNetV2~\citep{recht2019imagenet}, ImageNet-Sketch~\citep{wang2019learning}, ImageNet-A~\citep{hendrycks2021natural} and ImageNet-R~\citep{hendrycks2021many} with different types of domain shift. The compared methods include ProText~\citep{Khattak2024ProText} and ArGue-N~\citep{10657279} that use LLMs to distill language priors into the learned prompts, as well as more related regularization methods OGEN~\citep{zang2024overcoming} and CLAP~\citep{lavoie2024modeling}. OGEN regularizes the prediction probabilities with an improved Mean Teacher, while CLAP regularizes the class prototypes (\ie,~class-wise feature means) for linear probing.

Table~\ref{tb:Prompt_learning_2024} shows that our structure-wise feature regularization method Proxy-FDA outperforms OGEN and CLAP in all metrics under the considered settings. Proxy-FDA achieves particularly large gains in generalization performance on the new classes or new domains, maximizing the positive forward transfer with higher $\Delta_{\text{New}}$. When compared to ProText and ArGue-N using external LLMs, our approach is LLM-free but achieves on-par or even better performance for both prompt-tuning and OOD generalization.

\paragraph{Continual fine-tuning.} Table~\ref{tb:continual_fine_tuning} in the main paper compares our method with robust fine-tuning methods in the 3-task setting. In the same setting, Table~\ref{tb:continual_baseline} compares our method with 5 classic continual learning methods: LwF~\citep{Li17learning}, LFL~\citep{Jung16}, iCaRL~\citep{rebuffi2017icarl}, Distillation + Retrospection (D+R)~\citep{hou2018lifelong} and ZSCL~\citep{zheng2022preventing}.

Table~\ref{tb:continual_SplitImageNetR} compares our method with recent continual learning methods on the class-incremental learning benchmark Split ImageNet-R. This benchmark divides the 200 classes from ImageNet-R into 10 tasks with 20 classes per task. The compared methods include LDIFS as well as L2P~\citep{wang2022learning}, DualPrompt~\citep{wang2022dualprompt}, CODA-Prompt~\citep{Smith_2023_CVPR}, Continual-CLIP~\citep{thengane2022} and SLCA~\citep{zhang2023slca}. All methods use the same training (24,000) and testing (6,000) images. To further ensure fair comparisons, we follow the widely-adopted implementation: fine-tuning for 50 epochs using the Adam optimizer with $\beta_{1}=0.9$ and $\beta_{2}=0.999$. The initial learning rate is $1e^{-4}$, and we use a cosine learning rate
scheduler as in~\citep{mukhoti2024finetuning}.

In both Table~\ref{tb:continual_baseline} and~\ref{tb:continual_SplitImageNetR}, our (Proxy-)FDA method outperforms all other methods in preventing forgetting. At the same time, (Proxy-)FDA is able to achieve the best fine-tuning performance.

\section{Applications Beyond Classification}
\label{sec:appendix_more_tasks}

\paragraph{Fine-tuning for image captioning \& VQA.} Here we test if our (Proxy-)FDA method can address the forgetting issue for fine-tuning tasks beyond classification. Specifically, we consider the foundation model CLIP and fine-tune for two vision-language tasks: image captioning (COCO~\citep{LinMBHPRDZ14} and NoCaps~\citep{agrawal2019nocaps} datasets) and Visual Question-Answering (VQA2 dataset~\citep{balanced_vqa_v2}). The baseline approach that enables CLIP to perform such vision-language understanding and generation tasks is LiMBeR~\citep{merullo2023linearly}. LiMBeR maps the CLIP image features to the text space of a generative language model, using only a linear projection that aligns the image and text spaces. As a result, although the image encoder and language model are both frozen, LiMBeR allows CLIP to flexibly caption an image or perform some task relating to it.

For the ease of comparisons, we follow LiMBeR to use the same language model and image encoder (RN50x16) of CLIP. Starting with LiMBeR, we perform fine-tuning on COCO captions, and then benchmark the fine-tuning performance of COCO captioning as well as concept forgetting. We choose to measure forgetting in terms of the performance change between fine-tuned model and LiMBeR on two types of tasks: image captioning on a different dataset NoCaps, and VQA on VQA2. For efficient fine-tuning with only 5 captions per COCO image, we use the method of Visual Prompt Tuning (VPT)~\citep{jia2022visual} with CLIP and the language model kept frozen.

To evaluate image captioning performance, we report results of CIDEr-D~\citep{VedantamZP15}, CLIPScore, and Ref-CLIPScore~\citep{clipscore}. While for VQA, the model is prompted using the ``[image] Q: [q] A:'' format. The generated answer is truncated to the length of the longest ground truth answer. As the evaluation metric of VQA under the few-shot setting, accuracy is reported for every K-shot.

Table~\ref{tb:captioning_vqa} shows that VPT-based prompt tuning on COCO leads to forgetting on other tasks,~\eg,~$\Delta_{\text{Ref-S}}$ is negative on NoCaps captioning. On the other hand, LDIFS and our (Proxy-)FDA methods prove effective in regularizing the tuning process, all achieving positive forward transfer in all metrics. Encouragingly, our (Proxy-)FDA is better than LDIFS at promoting positive forward transfer, while maintaining competitive prompt tuning performance on COCO at the same time. 

\paragraph{Knowledge distillation.} As metioned in the Related Work section, the high-level idea of our method resembles Knowledge Distillation (KD), epseically those relational KD methods that distill feature relations between models.

Table~\ref{tb:knowledge_distillation} shows our method is directly applicable to KD and quite performant. We follow the standard KD settings in~\citep{zheng2024knowledge}, and test teacher-student pairs using the same or different architectures of ResNet~\citep{he2016deep} and MobileNet~\citep{howard2017mobilenet} on ImageNet. We compare with state-of-the-art \textit{logits matching} methods KD++~\citep{wang2023KD}, DIST~\citep{huang2022knowledge} and WTTM~\citep{zheng2024knowledge}. Note DIST can be viewed as a relational KD method at the logit level. We further compare with KD methods that \textit{match feature relations} in form of kNNs (CNA~\citep{zhu2022CNA}) and feature similarities (ITRD~\citep{miles2022itrd}). CNA and ITRD are more related to our FDA method, but FDA differs in that both neighbor indices and similarities are distilled in the feature space. We see from the table that FDA consistently outperforms CNA and ITRD, and is competitive or better than logits-based DIST. Our proxy learning further improves performance, and Proxy-FDA is on par with the best prior work WTTM.

\newpage

\begin{table*}[t!]
\centering
\caption{\textbf{Few-shot prompt tuning in the base-to-new class generalization setting} (16 shots per class). $\mathcal{A}_{\text{H}}$ denotes the Harmonic mean of $\mathcal{A}_{\text{Base}}$ and $\mathcal{A}_{\text{New}}$. $\Delta_{\text{New}}$ denotes the change in $\mathcal{A}_{\text{New}}$ between pre-trained and prompt-tuned CLIP models. Higher $\Delta_{\text{New}}$ shows lower level of concept forgetting on the new class split of the considered dataset.}
\label{tb:base2new_full}
\tablestyle{+1pt}{1.1}
\addtolength{\tabcolsep}{+3pt}
\resizebox{0.9\columnwidth}{!}{
\begin{tabular}{lc |cc |cc |cc |cc |cc |cc }
\toprule
 & \multicolumn{1}{c@{}}{} & \multicolumn{8}{c}{\textbf{Prompt tuning without regularization}} & \multicolumn{4}{c}{\textbf{Regularization-based}} \\ \cmidrule(lr){3-10} \cmidrule(lr){11-14}
 & \multicolumn{1}{c@{}}{} & \multicolumn{2}{c}{CoOp} & \multicolumn{2}{c}{CoCoOp} & 
\multicolumn{2}{c}{VPT} & \multicolumn{2}{c}{MaPLe} & \multicolumn{2}{c}{CLIPood} & \multicolumn{2}{c}{PromptSRC} \\
\cmidrule(lr){3-4} \cmidrule(lr){5-6} \cmidrule(lr){7-8} \cmidrule(lr){9-10} \cmidrule(lr){11-12} \cmidrule(lr){13-14}
~ & \multicolumn{1}{c@{}}{\textbf{+Proxy-FDA}} & \xmark & \multicolumn{1}{c@{}}{\cmark} & \xmark & \multicolumn{1}{c@{}}{\cmark} & \xmark & \multicolumn{1}{c@{}}{\cmark} & \xmark & \multicolumn{1}{c@{}}{\cmark} & \xmark & \multicolumn{1}{c@{}}{\cmark} & \xmark & \cmark \\
\midrule
\multirow{4}{*}{\shortstack[l]{\textbf{Avg across}\\ \textbf{11 datasets}}} & $\mathcal{A}_{\text{Base}}$ & 82.69 &  \textbf{83.16}  &  \textbf{80.47}  &  80.36  &  \textbf{81.61}  &  81.55  &  82.28  &  \textbf{82.74}  &  83.91   & \textbf{84.33}  &  84.26  &  \textbf{84.47}  \\
& $\mathcal{A}_{\text{New}}$ & 63.22 & \textbf{73.67}  &  71.69  &  \textbf{76.44}  &  69.61  &  \textbf{73.89}  &  75.14   & \textbf{77.13}  &  74.50  &  \textbf{76.54}  &  76.10  &  \textbf{77.45}  \\
& $\Delta_{\text{New}}\uparrow$ & -10.99 &\textbf{-0.55} &-2.53 &\textbf{2.22}  &-4.61 &\textbf{-0.33} &0.92  &\textbf{2.91} & 0.28 & \textbf{2.33} & 1.88 & \textbf{3.23} \\
& $\mathcal{A}_{\text{H}}$ & 71.66  &  \textbf{78.13}  &  75.83  &  \textbf{78.35}  &  75.14  &  \textbf{77.53}  &  78.55  &  \textbf{79.84}  &  78.93  &  \textbf{80.25}  &  79.97  &  \textbf{80.81}  \\
\midrule
\multirow{4}{*}{ImageNet} & $\mathcal{A}_{\text{Base}}$ & 76.47 & 76.22  &  75.98  &  76.95  &  75.96   & 75.26  &  76.66  &  77.35  &  77.50  &  78.47  &  77.60  &  77.81  \\
& $\mathcal{A}_{\text{New}}$ & 67.88 & 72.97  &  70.43  &  73.48  &  67.32  &  71.25  &  70.54  &  71.51 &   70.30  &  72.07  &  70.73 &   71.55  \\
& $\Delta_{\text{New}}\uparrow$ & -0.26 & 4.83 & 2.29 & 5.34 & -0.82 &3.11 & 2.40 &  3.37 &  2.16 & 3.93 & 2.59 & 3.41 \\
& $\mathcal{A}_{\text{H}}$ & 71.92  &  74.56  &  73.10  &  75.17  &  71.38 &   73.20  &  73.47  &  74.32  &  73.72  &  75.13  &  74.01  &  74.55  \\
\midrule
\multirow{4}{*}{Caltech101} & $\mathcal{A}_{\text{Base}}$ & 98.00 &  96.84  &  97.96  &  97.21  &  97.50  &  96.14  &  97.74  &  98.71  &  98.70  &  99.08  &  98.10  &  98.49  \\
& $\mathcal{A}_{\text{New}}$ & 89.81 & 97.45  &  93.81  &  97.15 &   94.10   & 95.93  &  94.36  &  95.42  &  94.60  &  95.01  &  94.03  &  95.34  \\
& $\Delta_{\text{New}}\uparrow$ & -4.19 & 3.45 & -0.19 &3.15 & 0.10 &  1.93 & 0.36 & 1.42 & 0.60  & 1.01 & 0.03 & 1.34\\
& $\mathcal{A}_{\text{H}}$ & 93.73  &  97.14  &  95.84  &  97.18  &  95.77  &  96.03  &  96.02  &  97.04  &  96.61 &   97.00  &  96.02  &  96.89  \\
\midrule
\multirow{4}{*}{OxfordPets} & $\mathcal{A}_{\text{Base}}$ & 93.67 &  95.01  &  95.20  &  96.96 &   96.05  &  95.32  &  95.43  &  95.42  &  95.70  &  97.63  &  95.33   & 96.31  \\
& $\mathcal{A}_{\text{New}}$ & 95.29 & 98.97  &  97.69  &  98.64 &   95.84  &  98.42  &  97.76  &  98.09 &   96.40  &  98.21  &  97.30  &  98.09  \\
& $\Delta_{\text{New}}\uparrow$ & -1.97 & 1.71 & 0.43 & 1.38 & -1.42 &1.16 & 0.50  & 0.83 & -0.86 &0.95 & 0.04 & 0.83\\
& $\mathcal{A}_{\text{H}}$ & 94.47  &  96.95  &  96.43  &  97.79  &  95.94  &  96.85  &  96.58   & 96.74  &  96.05  &  97.92  &  96.30 &   97.19  \\
\midrule
\multirow{4}{*}{\shortstack[l]{Stanford \\ Cars}} & $\mathcal{A}_{\text{Base}}$ & 78.12 &  78.33 &   70.49  &  69.53  &  75.00  &  74.16  &  72.94  &  74.01  &  78.60 &   78.07  &  78.27  &  77.95  \\
& $\mathcal{A}_{\text{New}}$ & 60.40 & 69.87  &  73.59   & 78.95  &  63.45   & 72.17   & 74.00  &  75.15  &  73.50 &   76.12  &  74.97   & 75.75  \\
& $\Delta_{\text{New}}\uparrow$ & -14.49 &-5.02 &-1.30 & 4.06 & -11.44 &  -2.72& -0.89 &0.26 & -1.39& 1.23 & 0.08&  0.86 \\
& $\mathcal{A}_{\text{H}}$ & 68.13  &  73.86 &   72.01  &  73.94  &  68.74   & 73.15  &  73.47  &  74.58  &  75.96  &  77.08  &  76.58  &  76.83  \\
\midrule
\multirow{4}{*}{Flowers102} & $\mathcal{A}_{\text{Base}}$ & 97.60 &  97.21  &  94.87  &  94.52  &  96.89 &   97.11  &  95.92 &   96.85 &   93.50  &  97.91 &   98.07  &  97.69  \\
& $\mathcal{A}_{\text{New}}$ & 59.67 & 72.36  &  71.75  &  77.54  &  70.02  &  73.49  &  72.46  &  75.59  &  74.50  &  76.59 &   76.50  &  78.49  \\
& $\Delta_{\text{New}}\uparrow$ & -18.13 &-5.44 &-6.05 &-0.26 &-7.78 &-4.31 &-5.34& -2.21 &-3.30 & -1.21& -1.30 & 0.69 \\
& $\mathcal{A}_{\text{H}}$ & 74.06  &  82.96  &  81.71  &  85.19  &  81.29  &  83.66  &  82.56 &   84.91 &   82.93  &  85.95 &   85.95 &   87.04  \\
\midrule
\multirow{4}{*}{Food101} & $\mathcal{A}_{\text{Base}}$ & 88.33  & 88.59  &  90.70  &  91.33   & 88.88  &  90.35 &   90.71  &  91.40   & 90.70 &   92.94  &  90.67  &  91.07  \\
& $\mathcal{A}_{\text{New}}$ & 82.26 & 90.12  &  91.29  &  94.79 &   88.95  &  92.27 &   92.05  &  93.12  &  91.70  &  92.76 &   91.53  &  92.25  \\
& $\Delta_{\text{New}}\uparrow$ & -8.96 & -1.10 & 0.07 & 3.57 & -2.27 &1.05 & 0.83 & 1.90 &  0.48 & 1.54 & 0.31 & 1.03 \\
& $\mathcal{A}_{\text{H}}$ & 85.19  &  89.35  &  90.99  &  93.03  &  88.91  &  91.30  &  91.38  &  92.25  &  91.20  &  92.85 &   91.10 &   91.66  \\
\midrule
\multirow{4}{*}{\shortstack[l]{FGVC \\ Aircraft}} & $\mathcal{A}_{\text{Base}}$ & 40.44 & 41.24 &   33.41  &  35.12  &  38.33 &   38.75  &  37.44  &  37.41  &  43.30  &  42.26  &  42.73  &  41.63  \\
& $\mathcal{A}_{\text{New}}$ & 22.30 & 33.83  &  23.71  &  36.36 &   25.27  &  31.36  &  35.61 &   37.79  &  37.20  &  37.54  &  37.87 &   40.61  \\
& $\Delta_{\text{New}}\uparrow$ & -13.99 &-2.46 &-12.58 &  0.07 & -11.02 &  -4.93 &-0.68 &1.50  & 0.91 & 1.25 & 1.58 & 4.32 \\
& $\mathcal{A}_{\text{H}}$ & 28.75  &  37.17  &  27.74  &  35.73  &  30.46  &  34.67  &  36.50  &  37.60  &  40.02   & 39.76 &   40.15  &  41.11  \\
\midrule
\multirow{4}{*}{SUN397} & $\mathcal{A}_{\text{Base}}$ & 80.60  &  80.63  &  79.74  &  80.36  &  80.27  &  79.54  &  80.82 &   81.24  &  81.00  &  83.04 &   82.67  &  82.71  \\
& $\mathcal{A}_{\text{New}}$ & 65.89 & 72.11  &  76.86  &  78.97  &  74.36  &  76.11  &  78.70  &  82.15  &  79.30  &  79.92  &  78.47  &  79.73  \\
& $\Delta_{\text{New}}\uparrow$ & -9.46 & -3.24 &1.51 & 3.62 & -0.99 &0.76 & 3.35 & 6.80  & 3.95 & 4.57 & 3.12 & 4.38 \\
& $\mathcal{A}_{\text{H}}$ & 72.51  &  76.13  &  78.27   & 79.66  &  77.20 &   77.79  &  79.75  &  81.69   & 80.14 &   81.45  &  80.52 &   81.19  \\
\midrule
\multirow{4}{*}{DTD} & $\mathcal{A}_{\text{Base}}$ & 79.44  &  79.51  &  77.01  &  75.92  &  77.08  &  76.68 &   80.36  &  80.05  &  80.80 &   80.14  &  83.37  &  84.04  \\
& $\mathcal{A}_{\text{New}}$ & 41.18 & 54.24  &  56.00  &  59.84  &  53.62  &  59.97  &  59.18  &  63.13 &   58.60  &  63.32  &  62.97  &  63.06  \\
& $\Delta_{\text{New}}\uparrow$ & -18.72 &-5.66& -3.90 & -0.06& -6.28 &0.07 & -0.72 &3.23 & -1.3 & 3.42 & 3.07 & 3.16 \\
& $\mathcal{A}_{\text{H}}$ & 54.24  &  64.49  &  64.85  &  66.93   & 63.24  &  67.30  &  68.16  &  70.59  &  67.93  &  70.74  &  71.75  &  72.05  \\
\midrule
\multirow{4}{*}{EuroSAT} & $\mathcal{A}_{\text{Base}}$ & 92.19 &  91.98  &  87.49  &  81.24  &  91.67  &  90.42  &  94.07  &  94.27  &  97.50  &  92.18  &  92.90  &  93.66  \\
& $\mathcal{A}_{\text{New}}$ & 54.74 & 78.29  &  60.04  &  66.87  &  58.31  &  67.02  &  73.23  &  75.11  &  64.10  &  71.01  &  73.90  &  77.12  \\
& $\Delta_{\text{New}}\uparrow$ & -9.31 & 14.24& -4.01 &2.82 & -5.74 &2.97 & 9.18 & 11.06 &0.05 & 6.96 & 9.85 & 13.07 \\
& $\mathcal{A}_{\text{H}}$ & 68.69  &  84.58 &   71.21 &   73.36  &  71.28  &  76.98   & 82.35 &   83.61  &  77.35 &   80.22 &   82.32  &  84.59  \\
\midrule
\multirow{4}{*}{UCF101} & $\mathcal{A}_{\text{Base}}$ & 84.69  &  89.15 &   82.33   & 84.86  &  80.07 &   83.37  &  83.00  &  83.43  &  85.70  &  85.95  &  87.10 &   87.79  \\
& $\mathcal{A}_{\text{New}}$ & 56.05 & 70.16  &  73.45  &  78.23   & 74.50   & 74.77  &  78.66  &  81.40  &  79.30  &  79.44  &  78.80  &  79.95  \\
& $\Delta_{\text{New}}\uparrow$ & -21.45 &-7.34 &-4.05& 0.73&  -3.00 &-2.73 &1.16 & 3.90  & 1.80  & 1.94 & 1.30  & 2.45 \\
& $\mathcal{A}_{\text{H}}$ & 67.46  &  78.52  &  77.64 &   81.41  &  77.18  &  78.84  &  80.77  &  82.40  &  82.38  &  82.57  &  82.74  &  83.69  \\
\bottomrule
\end{tabular}
}
\end{table*}

\begin{table*}[t!]
\centering
\caption{\textbf{Few-shot cross-dataset generalization} where CLIP is prompt-tuned on the source dataset ImageNet (16 shots per class) and tested on both ImageNet and 10 target datasets. We compare the test set accuracy $\mathcal{A}$ and the accuracy change $\Delta_{\mathcal{A}}$ (higher is better) between pre-trained and prompt-tuned models to quantify generalization and concept forgetting on each target dataset.}
\label{tb:cross_dataset}
\tablestyle{+5pt}{0.9}
\addtolength{\tabcolsep}{+3pt}
\resizebox{0.75\columnwidth}{!}{
\begin{tabular}{c|lc |cc |cc |cc }
\toprule
\multicolumn{1}{c@{}}{}& & \multicolumn{1}{c@{}}{} & \multicolumn{2}{c}{CoOp} & \multicolumn{2}{c}{CoCoOp} & \multicolumn{2}{c}{PromptSRC} \\
\cmidrule(lr){4-5} \cmidrule(lr){6-7} \cmidrule(lr){8-9}
\multicolumn{1}{c@{}}{}& & \multicolumn{1}{c@{}}{\textbf{+Proxy-FDA}} & \xmark & \multicolumn{1}{c@{}}{\cmark} & \xmark & \multicolumn{1}{c@{}}{\cmark} & \xmark & \multicolumn{1}{c@{}}{\cmark}\\
\midrule
\multirow{2}{*}{\textbf{Source}} & \multirow{2}{*}{ImageNet} & $\mathcal{A}$ & \textbf{71.51}  & 71.36  &  71.02  &  \textbf{71.24}  &  71.27  &  \textbf{71.32} \\
& & $\Delta_{\mathcal{A}}\uparrow$ & \textbf{4.78}  & 4.63 & 4.29 & \textbf{4.51} & 4.54 & \textbf{4.59} \\ \midrule
\multirow{22}{*}{\textbf{Target}} & \multirow{2}{*}{\shortstack[l]{\textbf{Avg across}\\ \textbf{10 datasets}}} & $\mathcal{A}$ & 63.88   &\textbf{66.09}  &  65.74 &   \textbf{66.48}  &  65.81  &  \textbf{66.86} \\
& & $\Delta_{\mathcal{A}}\uparrow$ & -1.20 & \textbf{1.01} & 0.66 & \textbf{1.40} & 0.72 & \textbf{1.78}  \\ \cmidrule(lr){2-9}
& \multirow{2}{*}{Caltech101} & $\mathcal{A}$ & 93.70  &  94.35   & 94.43  &  94.51  &  93.60  &  94.42   \\
& & $\Delta_{\mathcal{A}}\uparrow$ & 0.76 &  1.41 & 1.49 & 1.57 & 0.66 & 1.48  \\ \cmidrule(lr){2-9}
& \multirow{2}{*}{OxfordPets} & $\mathcal{A}$ & 89.14  &  90.53  &  90.14   & 90.62  &  90.25  &  90.78   \\
& & $\Delta_{\mathcal{A}}\uparrow$ & -0.07 & 1.32 & 0.93&  1.41 & 1.04 & 1.57 \\ \cmidrule(lr){2-9}
& \multirow{2}{*}{\shortstack[l]{Stanford \\ Cars}} & $\mathcal{A}$ & 64.51 &  66.18  &  65.32   & 66.22  &  65.70  &  66.55   \\
& & $\Delta_{\mathcal{A}}\uparrow$ & -0.81 & 0.86 & 0.00 & 0.90 &  0.38 & 1.23 \\ \cmidrule(lr){2-9}
& \multirow{2}{*}{Flowers102} & $\mathcal{A}$ & 68.71 & 71.54  &  71.88  &  72.32   & 70.25  &  72.04   \\
& & $\Delta_{\mathcal{A}}\uparrow$ & -2.63 & 0.20 &  0.54 & 0.98 & -1.09 & 0.70 \\ \cmidrule(lr){2-9}
& \multirow{2}{*}{Food101} & $\mathcal{A}$ & 85.30  &  86.86   & 86.06  &  86.91  &  86.15  &  87.38   \\
& & $\Delta_{\mathcal{A}}\uparrow$ & -0.76 & 0.80  & 0.00 & 0.85 & 0.09 & 1.32 \\ \cmidrule(lr){2-9}
& \multirow{2}{*}{\shortstack[l]{FGVC \\ Aircraft}} & $\mathcal{A}$ & 18.47  & 22.09  &  22.94  &  23.49  &  23.90  &  24.79   \\
& & $\Delta_{\mathcal{A}}\uparrow$ & -6.25&  -2.63& -1.78& -1.23& -0.82& 0.07  \\ \cmidrule(lr){2-9}
& \multirow{2}{*}{SUN397} & $\mathcal{A}$ & 64.15 & 66.12  &  67.36  &  67.62  &  67.10 &   67.53   \\
& & $\Delta_{\mathcal{A}}\uparrow$ & 1.65  & 3.62 & 4.86 & 5.12 & 4.60  & 5.03 \\ \cmidrule(lr){2-9}
& \multirow{2}{*}{DTD} & $\mathcal{A}$ & 41.92 & 45.13 &   45.73   & 46.15  &  46.87  &  47.31   \\
& & $\Delta_{\mathcal{A}}\uparrow$ & -2.47 & 0.74 & 1.34  &1.76 & 2.48 & 2.92 \\ \cmidrule(lr){2-9}
& \multirow{2}{*}{EuroSAT} & $\mathcal{A}$ & 46.39  &  49.08 &   45.37  &  47.89  &  45.50  &  48.37   \\
& & $\Delta_{\mathcal{A}}\uparrow$ & -1.21 & 1.48 & -2.23 &0.29 & -2.10 & 0.77 \\ \cmidrule(lr){2-9}
& \multirow{2}{*}{UCF101} & $\mathcal{A}$ & 66.55 & 69.01 &   68.21  &  69.10  &  68.75 &   69.42   \\
& & $\Delta_{\mathcal{A}}\uparrow$ & -0.20  & 2.26 & 1.46 & 2.35 & 2.00 & 2.67 \\
\bottomrule
\end{tabular}
}
\end{table*}

\begin{table*}[t]
\caption{\textbf{Few-shot prompt tuning in both base-to-new class generalization and domain generalization settings.} Here we compare with more recent prompt tuning methods. Note both OGEN and our Proxy-FDA are plugged into the PromptSRC baseline. For fair comparison with CLAP, we obtain its base-to-new generalization results by re-running its official codes with the ViT-B/16 backbone used by all other methods. The domain generalization results of CLAP are directly extracted from the CLAP paper. $\mathcal{A}_{\text{H}}$ denotes the Harmonic mean of $\mathcal{A}_{\text{Base}}$ and $\mathcal{A}_{\text{New}}$.}
\label{tb:Prompt_learning_2024}
\begin{center}
\resizebox{0.9\linewidth}{!}{
\begin{tabular}{cl|ccccccccc}
\toprule
&\multicolumn{1}{l@{}}{} & \multicolumn{4}{c}{\textbf{Base-to-New Class Generalization}} & \multicolumn{5}{c}{\textbf{Domain Generalization}} \\
\cmidrule(lr){3-6} \cmidrule(lr){7-11}
&\multicolumn{1}{l@{}}{} & \multicolumn{4}{c}{Avg across 11 datasets} & \multicolumn{1}{c}{$\mathcal{A}_{\text{Source}}$} & \multicolumn{4}{c}{$\mathcal{A}_{\text{Target}}$} \\
\cmidrule(lr){3-6} \cmidrule(lr){7-7} \cmidrule(lr){8-11}
& \multicolumn{1}{l@{}}{} & $\mathcal{A}_{\text{Base}}$ & $\mathcal{A}_{\text{New}}$ & $\Delta_{\text{New}}\uparrow$ & $\mathcal{A}_{\text{H}}$ & ImageNet & -V2 & -Sketch & -A & -R \\
\midrule 
\textbf{Text Knowledge} & ProText & 72.95 & 76.98 & 2.76 & 74.91 & 70.22 & 63.54 & 49.45 & 51.47 & 77.35 \\
\textbf{from LLM} & ArGue-N & 83.77 & \textbf{78.74} & \textbf{4.52} & \textbf{81.18} & 71.84 & 65.02 & 49.25 & 51.47 & 76.96 \\ \midrule 
\multirow{3}{*}{\shortstack[c]{\textbf{Regularization} \\\textbf{method} }} & OGEN & 84.17 & 76.86 & 2.64 & 80.34 & 73.13 & 65.37 & 48.96 & 50.75 & 77.12 \\
& CLAP & 84.34 & 76.62 & 2.40 & 80.29 & 73.38 & 65.00 & 48.35 & 49.53 & 77.26 \\
& Proxy-FDA  & \textbf{84.47} & 77.45 & 3.23 & 80.81 & \textbf{73.44} & \textbf{65.79} & \textbf{49.83} & \textbf{51.54} & \textbf{77.45}\\
\bottomrule
\end{tabular}
}
\end{center}
\end{table*}

\clearpage
\newpage

\begin{table*}[!t]
\vskip 0.1in
\caption{\textbf{Continual fine-tuning: test accuracy $\mathcal{A}_{\text{LP}}$ and $\Delta_{\text{LP}}$ for models fine-tuned on three task sequences}.
The first 3 rows show performance on fine-tuned tasks and the 4th row shows performance averaged on 6 other datasets, comparing our method with 5 classic continual learning methods.}
\vskip -0.2in
\label{tb:continual_baseline}
\begin{center}
\resizebox{1.0\linewidth}{!}{
\begin{tabular}{ll|cc|cc|cc|cc|cc|cc|cc}
\toprule
Fine-tune & \multicolumn{1}{l@{}}{Evaluation} & \multicolumn{2}{c}{LwF} & \multicolumn{2}{c}{LFL} & \multicolumn{2}{c}{iCaRL} & \multicolumn{2}{c}{D+R} & \multicolumn{2}{c}{ZSCL} & \multicolumn{2}{c}{FDA (ours)} & \multicolumn{2}{c}{Proxy-FDA (ours)} \\
\cmidrule(lr){3-4} \cmidrule(lr){5-6} \cmidrule(lr){7-8} \cmidrule(lr){9-10} \cmidrule(lr){11-12} \cmidrule(lr){13-14} \cmidrule(lr){15-16}
dataset& \multicolumn{1}{l@{}}{dataset} & $\mathcal{A}_{\text{LP}}$ & \multicolumn{1}{c@{}}{$\Delta_{\text{LP}}\uparrow$}& $\mathcal{A}_{\text{LP}}$ & \multicolumn{1}{c@{}}{$\Delta_{\text{LP}}\uparrow$} & $\mathcal{A}_{\text{LP}}$ & \multicolumn{1}{c@{}}{$\Delta_{\text{LP}}\uparrow$} & $\mathcal{A}_{\text{LP}}$ & \multicolumn{1}{c@{}}{$\Delta_{\text{LP}}\uparrow$} & $\mathcal{A}_{\text{LP}}$ & \multicolumn{1}{c@{}}{$\Delta_{\text{LP}}\uparrow$} & $\mathcal{A}_{\text{LP}}$ & $\Delta_{\text{LP}} \uparrow$ & $\mathcal{A}_{\text{LP}}$ & $\Delta_{\text{LP}} \uparrow$\\
\midrule
\multirow{4}{*}{\shortstack[l]{SVHN$\rightarrow$\\CIFAR10$\rightarrow$\\RESISC45}} & SVHN & 90.48 &-3.81 &91.90 &-3.21 & 91.62 &-3.67 &93.30& -2.78 &92.70 &-3.23 &\textbf{96.77} &0.61 &96.72 &\textbf{0.93}\\
& CIFAR10  & 93.90 &-2.90 &94.88 &-2.32 &95.17 &-2.10 &95.41& -1.90 & 95.82& -1.60 &97.13 &0.57 &\textbf{97.29} &\textbf{1.02}\\
& RESISC45 & 94.22 &3.10 &93.90 &2.98 &93.72 &2.83 &94.94 &3.68 & 94.89& 3.62&95.22 &4.14 &\textbf{95.38} &\textbf{4.22}\\ \cmidrule(lr){2-16}
& Others & 80.73 &-4.20 &81.31 &-3.76 &80.78 &-4.11 &81.86 &-3.20 & 83.10 & -2.80 &\textbf{87.21} &0.76 &86.95 &\textbf{1.08} \\ \midrule
\multirow{4}{*}{\shortstack[l]{SVHN$\rightarrow$\\CIFAR100$\rightarrow$\\RESISC45}} & SVHN & 89.48 &-4.34 &90.29 &-4.08 & 90.97 &-4.31 &92.30& -3.23 & 91.81& -3.92 &96.18 &0.63 &\textbf{96.43} &\textbf{0.71}\\
& CIFAR100  & 83.24 &-3.25 &83.95 &-3.01 &84.06 &-3.13 &84.82& -2.60 & 85.07& -2.13 &\textbf{86.33} &0.72 &86.14 &\textbf{0.85}\\
& RESISC45 & 93.80 &3.21 &94.91 &3.62 &94.87 &3.54 &95.08 &3.71 & 94.96& 3.65&95.32 &3.95 &\textbf{95.46} &\textbf{4.01}\\ \cmidrule(lr){2-16}
& Others & 81.73 &-4.11 &82.04 &-3.80 &81.62 &-4.02 &82.17 &-3.43 & 82.86& -3.11 &89.02 &0.68 &\textbf{89.09} &\textbf{0.96} \\ \midrule
\multirow{4}{*}{\shortstack[l]{SVHN$\rightarrow$\\Cars$\rightarrow$\\RESISC45}} & SVHN & 91.43 &-3.64 &92.74 &-2.92 & 91.75 &-3.13 &92.86& -2.84 & 92.98& -2.72 &96.74 &0.79 &\textbf{96.91} &\textbf{0.94}\\
& Cars  & 81.69 &-2.79 &81.82 &-2.64 &81.70 &-2.80 &82.11& -2.12 & 82.68& -1.84 &\textbf{84.38} &1.14 &84.32 &\textbf{1.36}\\
& RESISC45 & 93.92 &3.34 &94.96 &3.55 &94.97 &3.58 &95.19 &3.72 & 95.04 & 3.63 &95.12 &3.92 &\textbf{95.23} &\textbf{4.07}\\ \cmidrule(lr){2-16}
& Others & 81.63 &-4.07 &82.24 &-3.60 &81.88 &-3.89 &82.73 &-3.12 & 83.10&-2.80 &89.54 &0.96 &\textbf{89.67} &\textbf{1.17} \\
\bottomrule
\end{tabular}
}
\end{center}
\end{table*}

\begin{table*}[!t]
\vskip 0.1in
\caption{\textbf{Continual fine-tuning:} comparing the average accuracy on Split ImageNet-R.}
\label{tb:continual_SplitImageNetR}
\begin{center}
\vskip -0.2in
\resizebox{1.0\linewidth}{!}{
\begin{tabular}{cccccccc}
\toprule
L2P & DualPrompt & CODA-Prompt & Continual-CLIP & SLCA & LDIFS & FDA (ours) & Proxy-FDA (ours) \\
\midrule
74.60$\pm$1.21 & 77.24$\pm$1.27 & 78.13$\pm$1.18 & 76.23$\pm$1.18 & 81.22$\pm$1.23 & 83.62$\pm$1.16 & 85.97$\pm$1.05 & \textbf{86.71$\pm$1.24} \\
\bottomrule
\end{tabular}
}
\end{center}
\end{table*}

\begin{table*}[!t]
\caption{\textbf{Prompt tuning for image captioning and VQA}. The CLIP model with LiMBeR projection is prompt-tuned on COCO dataset, and the fine-tuning performance for COCO captioning is reported in three metrics: CIDEr-D, CLIPScore, and Ref-CLIPScore. While forgetting is benchmarked in terms of the performance change between prompt-tuned and original models on two different tasks: captioning on NoCaps (in $\Delta_{\text{CIDEr-D}}$, $\Delta_{\text{CLIP-S}}$, $\Delta_{\text{Ref-S}}$), and VQA on VQA2 (in accuracy change $\Delta_{\mathcal{A}}$). Higher performance change indicates lower forgetting or even positive forward transfer ($\Delta>0$).}
\label{tb:captioning_vqa}
\begin{center}
\vskip -0.2in
\resizebox{1.0\linewidth}{!}{
\begin{tabular}{lccc ccc cccc}
\toprule
 &\multicolumn{6}{c}{\textbf{Image Captioning}} & \multicolumn{4}{c}{\textbf{VQA2 K-shots}} \\
 \cmidrule(lr){2-7}
 \cmidrule(lr){8-11}
 & \multicolumn{3}{c}{COCO} & \multicolumn{3}{c}{NoCaps} & 0 & 1 & 2 & 4  \\
\cmidrule(lr){2-4}
\cmidrule(lr){5-7}
\cmidrule(lr){8-11}
 &  CIDEr-D & CLIP-S & Ref-S & $\Delta_{\text{CIDEr-D}}\uparrow$ & $\Delta_{\text{CLIP-S}}\uparrow$ & $\Delta_{\text{Ref-S}}\uparrow$ & $\Delta_{\mathcal{A}}\uparrow$ & $\Delta_{\mathcal{A}}\uparrow$& $\Delta_{\mathcal{A}}\uparrow$& $\Delta_{\mathcal{A}}\uparrow$ \\ \midrule
VPT~\citep{jia2022visual} & \textbf{57.1} & 79.6 & 82.8 & 0.6 & 1.2 & -0.3 & 1.1 & 0.7 & 0.8 & 1.9\\
VPT+LDIFS & 56.8 & 80.3 & 82.4 & 1.5 & 1.8 & 0.6 & 2.1 & 1.3 & 1.6 & 3.2\\
VPT+FDA (ours) & 56.2 & 80.7 & \textbf{83.4} & 2.2 & 2.3 & 1.4 & 2.6 & 1.5 & 2.1 & 3.9\\
VPT+Proxy-FDA (ours) & 56.6 & \textbf{81.1} & 83.2 & \textbf{2.6} & \textbf{2.5} & \textbf{1.7} & \textbf{2.7} & \textbf{1.9} & \textbf{2.4} & \textbf{4.4}\\
\bottomrule
\end{tabular}
}
\end{center}
\end{table*}

\begin{table*}[!t]
\caption{\textbf{Knowledge distillation:} comparing the top-1 accuracy on ImageNet.}
\label{tb:knowledge_distillation}
\begin{center}
\vskip -0.2in
\resizebox{1.0\linewidth}{!}{
\begin{tabular}{cc|ccccccc}
\toprule 
 & \multicolumn{1}{c@{}}{} & \multicolumn{3}{c}{\textbf{Logits-based}} & \multicolumn{4}{c}{\textbf{Feature-based}} \\
 \cmidrule(lr){3-5} \cmidrule(lr){6-9}
Teacher & \multicolumn{1}{c@{}}{Student} & KD++ & DIST & WTTM & CNA & ITRD & FDA (ours) & Proxy-FDA (ours) \\
\midrule
ResNet-34 (73.31) & ResNet-18 (69.76) & 71.98 & 72.07 & \textbf{72.19} & 71.38 & 71.68 & 72.02 & 72.17\\
ResNet-50 (76.16) & MobileNet (68.87) & 72.77 & 73.24 & 73.09 & 72.39 & - & 73.31 & \textbf{73.45}\\
\bottomrule
\end{tabular}
}
\end{center}
\end{table*}


\end{document}